%% file: main.tex
\pdfoutput=1

\documentclass[11pt]{article}
\usepackage[preprint]{acl}

\usepackage{times}
\usepackage{latexsym}

\usepackage[T1]{fontenc}

\usepackage[utf8]{inputenc}

\usepackage{microtype}

\usepackage{inconsolata}

\usepackage{graphicx}

\usepackage{outlines}
\usepackage{xspace}
\usepackage{amsfonts}
\usepackage{amsthm}
\usepackage{amsmath}
\usepackage{amssymb}
\usepackage{bbold}
\usepackage{cleveref}
\usepackage{bbm}
\usepackage{caption}
\usepackage{subcaption}
\usepackage{xcolor}
\usepackage{thm-restate}
\usepackage{booktabs}
\usepackage{enumitem}
\usepackage{multirow}
\usepackage{hyperref} 
\usepackage{scalerel}
\usepackage{tabularx}
\usepackage{tikz}
\usetikzlibrary{patterns.meta}

\input{macros}
\input{venn_diagram_macros}

\title{What Do Prosody and Text Convey? Characterizing How Meaningful Information is Distributed Across Multiple Channels}

\author{
    Aditya Yadavalli \\
    UC San Diego \\
    \href{mailto:a1yadavalli@ucsd.edu}{\texttt{a1yadavalli@ucsd.edu}}\\\And
    Tiago Pimentel \\
    ETH Zürich \\
    \href{mailto:tiago.pimentel@inf.ethz.ch}{\texttt{tiago.pimentel@inf.ethz.ch}} \\\AND
    Tamar I. Regev \\
    MIT \\
    \href{mailto:tamarr@mit.edu}{\texttt{tamarr@mit.edu}} \\\And
    Ethan Gotlieb Wilcox \\
    Georgetown University \\
    \href{mailto:ethan.wilcox@georgetown.edu}{\texttt{ethan.wilcox@georgetown.edu}}\\\And
    Alex Warstadt \\
    UC San Diego \\ 
    \href{mailto:awarstadt@ucsd.edu}{\texttt{awarstadt@ucsd.edu}}
    }

\begin{document}
\maketitle

\begin{abstract}
Prosody -- the melody of speech -- conveys critical information often not captured by the words or text of a message.
In this paper, we propose an information-theoretic approach to quantify how much information is expressed by prosody alone and not by text, and crucially, what that information is about.
Our approach applies large speech and language models to estimate the mutual information between a particular dimension of an utterance's meaning (e.g., its emotion) and any of its communication channels (e.g., audio or text).
We then use this approach to quantify how much information is conveyed by audio and text about sarcasm, emotion, and questionhood, using speech from television and podcasts.
We find that for sarcasm and emotion the audio channel -- and by implication the prosodic channel -- transmits over an order of magnitude more information about these features than the text channel alone, at least when long-term context beyond the current sentence is unavailable.
For questionhood, prosody provides comparatively less additional information.
We conclude by outlining a program applying our approach to more dimensions of meaning, communication channels, and languages.
\end{abstract}

\section{Introduction}

Language is more than just the words and phrases we speak, sign, or write down.
Linguistic communication typically takes place over multiple modalities and channels, including facial expression, gesture, typography, and speech prosody \citep{aristotle1991rhetoric,kress2009multimodality,patel2023super}.
Prosody, in particular, encompasses the suprasegmental features of speech such as pitch, tempo, and loudness, and plays an essential role in communication \citep{hirschberg2002communication}.
In written communication, the audio channel containing the prosody is completely removed, leaving behind mainly the lexical content, often referred to as the ``segmental information'', but which we refer to as the \newterm{text}.\footnote{
However, the information conveyed by prosody is so critical to successful communication that humans have developed numerous typographical conventions to recover much of that information loss, including commas, question marks, italics, and emojis \citep{chafe1988punctuation,holtgraves2020emoji}.}
In this paper, we examine how much information about an utterance's meaning is contained in that audio, and how much information is lost when only the text is available.\looseness=-1

\begin{figure}
    \centering
    \input{venn_diagram}
    \caption{The informational relationship between linguistic channels -- Text \Text, Prosody \Prosody, Audio \Audio -- and a meaningful feature \Feat. Features are discrete properties of a message, such as its syntax or the speaker's affect.}
    \label{fig:overview}
\end{figure}
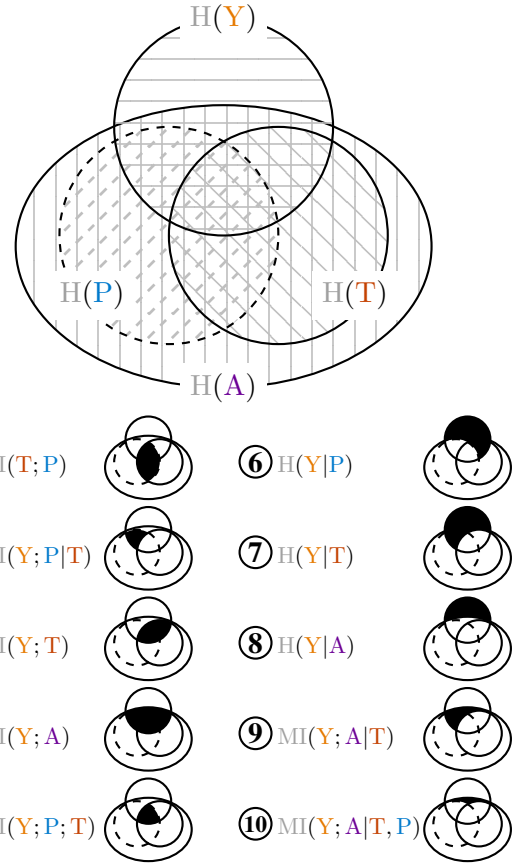

There is a growing trend of applying information theory to questions in linguistics and language processing \citep{levy2008expectationbased,piantadosi2011word,futrell2020lossy,williams2021relationships,socolof2022measuring}. 
Particularly relevant, some recent work has investigated how information is distributed across the prosodic and textual channels \citep{wolf2023redundancy,regev2025time,wilcox2025using}.
However, this prior work gives an incomplete picture:
Its focus has been entirely on using mutual information estimation to quantify the overall \emph{redundancy} between text and prosody.
While it is intriguing that text and prosody convey overlapping information, this methodology gives us no insight into how much \emph{unique} information prosody conveys and what that information is \emph{about}.

Here, we instead characterize how information about a particular meaningful feature (\Feat, e.g., whether an utterance is sarcastic) is distributed over distinct channels (prosody \Prosody, audio \Audio, and text \Text).
Formally, a feature \Feat's information is quantified by its entropy, $\entropy(\Feat)$.
As illustrated in \cref{fig:overview}, this entropy can be decomposed as the sum of several information-theoretic terms ($\entropy(\Feat) = \legendCircleThree + \legendCircleTwo + \legendCircleTen + \legendCircleEight$).
Thus, estimating these terms gives us a more complete picture of how information about \Feat is distributed across prosody and text.
Following \citet{pimentel2020information}, we estimate these information-theoretic terms by fine-tuning pretrained models to predict \Feat given one or more channels as input.\footnote{
Choosing \Feat to be a discrete variable, \mi estimation largely consists of training simple classifiers and measuring their cross-entropy loss on a dataset labeled for \Feat. When using \Audio as our channel, our classifiers are based on spoken language models (SLM); when using \Text, they are based on language models (LMs).}\looseness=-1

As case studies, we apply our method to two features which we predict are strongly associated with prosodic signals -- sarcasm and emotion -- and to one feature we predict to be less strongly associated with prosody due to strong textual cues -- questionhood. We find that the audio channel, including the prosody, conveys roughly an order of magnitude more information than text alone about the two first linguistic features -- at least when the length of textual context is limited to a single sentence -- but only 2.4 times as much for the latter feature.

More broadly, our approach introduces a framework that can be applied to many other research questions.
Prosody conveys crucial information about syntax, lexical identity, speech act, and discourse --  all of which can be investigated using our approach.
Furthermore, any communication channel -- be it the pitch contour, gesture, or emojis -- can be similarly explored, as long as a classifier can be trained to accept input from that channel.
Finally, languages differ significantly in the ways they use prosody \citep{hyman2008universals,wilcox2025using} as well as other communication channels. Our approach can give fine-grained insight into the diversity of strategies that can be used for distributing information across communication channels.

\section{Background}

\subsection{Prosody}

Prosody is the suprasegmental component of speech, including all the auditory features of the speech signal apart from those that signal to segmental features (i.e., phonemes). 
This encompasses pitch (fundamental frequency), tempo (duration and pauses), and loudness, but also more subtle spectral properties, such as harmonics-to-noise ratio or resonance changes affecting voice quality \cite{CHEANG2008366}. 
Prosodic cues can convey critical information, including the locations of word boundaries \citep{cutler2014prosody}, syntactic disambiguation \citep{pauker2011prosody, snedeker2003using}, information about lexical identity \citep{wilcox2025using}, grammatical marking \citep[e.g., whether or not an utterance is a question;][]{Cole2015ProsodyIC, hellbernd2016prosody}, and information structure in a sentence \cite{breen2010acoustic, roettger2019mapping}.
Speakers can also use prosody to convey paralinguistic information, such as affect \citep{fernandez2011affect}, intent, and signals that they are about to end their turn in a conversation \citep{cutler2018analysis}.
Importantly for this study, much of the information conveyed by prosody is absent in written language. 
\todo[cite superlinguistics if possible]

\subsection{Spoken Language Models}\label{sec:slm}
Transformers \cite{NIPS2017_3f5ee243} have made efficient processing of sequential data possible. 
While they were initially used for text processing \cite{devlin-etal-2019-bert, Radford2019LanguageMA}, they have also enabled many recent advancements in spoken language modeling.
The speech representations output by Transformer-based models pretrained on unlabeled audio data such as wav2vec \citep{schneider2019wav2vec,baevski2020wav2vec}, HuBERT \citep{hsu2021hubert}, and many others \cite[e.g.,][]{conneau2020unsupervised,babu2021xls,pratap2024scaling} have been successfully applied to tasks such as speech recognition, speaker identification, and affect classification with a limited amount of labeled training data. 
In particular, 
the Whisper models \cite{radford2022whisper} were trained on 680,000 hours of audio data across speech recognition and other speech-to-text tasks and, 
at the time of their release, they showed state-of-the-art performance on many of these tasks across accents, languages, acoustic conditions, and speaker variability, leading to its widespread adoption \cite{radford2022whisper, olatunji2023afrispeech, javed2023svarah, bhogale2023vistaar, talafha2023n}.\looseness=-1

These advances in spoken language modeling have significant implications for human-computer interaction. 
As Large Language Models (LLMs) are increasingly being used in user-facing applications, they require capabilities beyond text understanding -- particularly the ability to detect users' affect and intents from vocal cues. \tocite
While text-only models can infer affect only from orthography, audio-based models have access to crucial paralinguistic features like tone, pitch variations, and speech rate that often convey more reliable affective signals. 
This is why spoken language understanding, which integrates these audio capabilities with classical language understanding approaches, has emerged as a critical field bridging the gap between spoken language processing advancements and practical applications in conversational AI systems \cite{shon2022slue, 9746137, shon2024discreteslu, arora2024evaluation}. 
In addition to serving many practical purposes, the goal of our contribution is to demonstrate how speech models can also enable us to address fundamental scientific questions about language. 




\section{Information Content of Prosody}


Formally, our objective is to study how information is distributed among four random variables: \Text, a \defn{text}-valued random variable; \Prosody, \defn{prosody}-valued; \Audio, which is \defn{audio}-valued; and \Feat, which assumes values of a \defn{target feature} of interest (e.g., affect or sarcasm).
We represent each of these random variables as upper-case letters (e.g., \Text), and their instances with lower-case letters (e.g., \textvar).
We also use \Channel to denote any random variable representing a communication channel (or a combination thereof).
For our purposes, a text \textvar is an orthographic representation of a single utterance.
A prosody-valued instance \prosody is a joint representation of all the values of each prosodic feature across an entire utterance.
An audio-valued instance \audio represents the full waveform for an utterance.
Finally, a target feature \feat represents the value this feature takes in an utterance, for instance, whether the entire utterance is sarcastic.
All utterances we will consider contain a single sentence in English.

\subsection{Mutual Information Estimation}

The \newterm{mutual information} (\mi) between random variables \Channel and \Feat, with supports \Channelsupport and \Featsupport, is defined as:%
%
\begin{equation}
    \mi(\Channel;\Feat) = \sum_{\channel \in \Channelsupport, \feat \in \Featsupport} \prob(\channel, \feat) \log\frac{\prob(\channel, \feat)}{\prob(\channel)\prob(\feat)}.
\label{eq:mi_defn}
\end{equation} 
Mutual information is an information-theoretic measure that quantifies (e.g., in bits) how much the uncertainty about one random variable can be reduced when another random variable is observed \cite{Shannon1948AMT, shannon1951prediction}.
More informally, \mi is the amount of overlap in information between two variables (e.g., between sarcasm and prosody).

Due to the difficulty of estimating the joint distribution $\prob(\Channel,\Feat)$,
\citet{pimentel2020information} propose to analyze
mutual information as a difference of two entropies, given the following identity,
\begin{equation}
    \mi(\Channel;\Feat) = \entropy(\Feat) - \entropy(\Feat\mid\Channel),
\label{eq:mi_identity}
\end{equation} 
where these Shannon entropies \entropy \citep{Shannon1948AMT} are defined as follows:
\begin{align}
    \entropy(\Feat) &= -\sum_{\feat \in \Featsupport} \prob(\feat)\log\prob(\feat),
\label{eq:ent_defn} \\
    \entropy(\Feat\mid\Channel) &= -\sum_{\channel \in \Channelsupport,\feat \in \Featsupport} \prob(\channel,\feat)\log\prob(\feat\mid\channel).
\label{eq:cond_ent_defn}
\end{align} 
%



\citet{wolf2023redundancy} recently adopted this mutual information decomposition to measure $\mi(\Text;\Prosody)$, the redundancy between text \Text and prosody \Prosody.%
\footnote{\citeauthor{wolf2023redundancy} formulate \Prosody as ranging over the prosody of a single word. This difference does not affect our discussion.}
While quantifying this redundancy can provide some insight, it does not tell us what this information is actually \emph{about}.

\subsection{Our Approach} \label{sec:approach}

Here, we seek to quantify how much information \emph{about} a specific type of meaning is shared by text and prosody, and how much is conveyed by them separately.
The central idea
is that, rather than estimate the mutual information of prosody and text directly $\mi(\Text; \Prosody)$,
we can estimate their mutual information with some meaningful feature \Feat instead.\looseness=-1
%

Our approach is summarized in \cref{fig:overview}.
This diagram illustrates how different information-theoretic quantities of interest relate to one another.
As discussed above, the quantity measured by \citet{wolf2023redundancy}, $\mi(\Text;\Prosody)$, is \legendCircleOne.
We are interested in a larger set of quantities.
We ultimately aim to characterize the relationship of the text \Text, prosody \Prosody, and some meaningful feature \Feat.
To estimate these values, however, we train classifiers to predict \Feat from channel \Channel. 
While we can effectively train audio-based classifiers (using  existing audio encoder models, see Section \ref{sec:slm}), no such architecture exists for prosody (see Section \ref{sec:limitations} for further discussion).
We thus choose to estimate these terms using the full audio signal \Audio instead of \Prosody.

Fortunately, we can estimate many (but not all) of the relevant quantities for prosody by studying just the text and the full audio signal.
The information conveyed by audio $\entropy(\Audio)$ contains all of the information conveyed by text $\entropy(\Text)$ and prosody $\entropy(\Prosody)$.
Thus, \legendCircleFour $\mi(\Feat;\Audio)$ contains information about \Feat that can be obtained from either text, prosody, or some other aspect of the audio.
We estimate this quantity using the following identity
\begin{equation}
    \mi(\Feat;\Audio) = \entropy(\Feat) - \entropy(\Feat\mid\Audio).
\end{equation}
In the same way, we can estimate \legendCircleThree $\mi(\Feat;\Text)$ as 
\begin{equation}
    \mi(\Feat;\Text) = \entropy(\Feat) - \entropy(\Feat\mid\Text),
\end{equation}

Then, we can approximate \legendCircleTwo $\mi(\Feat;\Prosody\mid\Text)$ given the above {\mi}s.
This quantity tells us how much information prosody conveys about \Feat that is not conveyed through the text.
In practice, though, we compute a different conditional \mi,\footnote{Following the definition of \mi, technically $\mi(\Feat;\Audio\mid\Text)$ = $\mi(\Feat;\Audio, \Text) - \mi(\Feat;\Text)$. However, this can be simplified to \cref{eq:audio_text} under our assumption that \Audio fully determines \Text.}
\begin{align} \label{eq:audio_text}
     \mi(\Feat;\Audio\mid\Text) = \mi(\Feat;\Audio) - \mi(\Feat;\Text).
\end{align}
However, we argue that the quantity \legendCircleNine $\mi(\Feat;\Audio\mid\Text)$ is a good approximation for \legendCircleTwo $\mi(\Feat;\Prosody\mid\Text)$.
This is because
\legendCircleNine overestimates \legendCircleTwo only by the term \legendCircleTen $\mi(\Feat;\Audio\mid\Text,\Prosody)$, where
\legendCircleTen is all the information in the audio signal about \Feat that is not conveyed by either text or prosody.
We reason that this quantity is minimal: Most of the information in audio that does not fall under text or prosody is likely to be irrelevant information about background noise, accent, or other qualities of speaker identity.%
\footnote{This quantity may arguably include relevant content such as laughter, filler words, and small variations in phoneme production which may or may not be defined as prosodic.}
Thus, we can approximately isolate the unique contribution of the prosodic channel to communication about a particular kind of meaning, such as sarcasm, without requiring access to a specialized pretrained prosody model.\looseness=-1

\section{Estimating Mutual Information} \label{sec:estimation_methods}

Prior work \citep{pimentel2020information,williams2021relationships,wolf2023redundancy} has used neural network-based classifiers to estimate the mutual informations above.
In particular,
\citet{pimentel2020information} propose a method to estimate $\mi(\Channel;\Feat)$ given a dataset \dataset, composed of pairs $(\channel_i, \feat_i)$ sampled from the joint distribution $\prob(\Channel,\Feat)$.
We follow their approach here.\looseness=-1

Relying on the decomposition in \cref{eq:mi_identity},
we must estimate two entropies.
First, the unconditional entropy $\entropy(\Feat)$ is estimated from the dataset non-parametrically using a plug-in estimation: 
\begin{equation}
    \prob(\feat) \approx \frac{1}{|\dataset|}\sum_{i=1}^{|\dataset|} \indicator_{\feat}(\feat_i).
\label{eq:defn_plugin}
\end{equation} 
Second, the conditional entropy $\entropy(\Feat\mid\Channel)$ is estimated using a neural network with parameters $\theta$ optimized on a separate dataset \datasettrain to predict each $\feat_i$ from each $\channel_i$. 
This model learns a conditional distribution $\ptheta(\Feat \mid \Channel)$, which can be used to compute the cross-entropy $\xent(\Feat \mid \Channel)$ as follows:
\begin{equation}
    \xent(\Feat\mid\Channel) \approx -\frac{1}{|\datasettest|}\sum_{i=1}^{|\datasettest|} \log\ptheta(\feat_i\mid\channel_i).
\label{eq:xent}
\end{equation} 
As $\xent(\Feat\mid\Channel) \geq \entropy(\Feat\mid\Channel)$, we want to obtain the tightest upper bound on $\entropy(\Feat\mid\Channel)$ by training the best model as possible.
In our case, this means fine-tuning all the weights of a large foundation model to predict \Feat from $\Channel$.

\paragraph{Comparison to Prior Work Estimating \mi with Prosody.}

Our estimation technique largely consists of training simple neural network classifiers to predict meaningful features like sarcasm from some input channel like text or audio.
Prior information-theoretic approaches to studying prosody \citep{wolf2023redundancy,wilcox2025using,regev2025time} estimate $\mi(\Text;\Prosody)$ by decomposing it into $\entropy(\Prosody)$ and $\entropy(\Prosody\mid \Text)$, which requires predicting the continuous variable \Prosody.
By treating the discrete variable \Feat as the dependent variable, our method avoids two key pitfalls of prior work:

First, the entropy of continuous variables, known as \newterm{differential entropy}, can be negative and does not have an intuitive interpretation.
While the difference of $\entropy(\Prosody\mid\Text)$ and $\entropy(\Prosody)$ can still be interpreted as the \mi,%
\footnote{
In such cases where one measures the \mi of a continuous and a discrete variable, the equality in \cref{eq:mi_identity} requires additional definitions and assumptions (namely, the good mixed pair assumption; see \citealp{wolf2023redundancy} for details).
This caveat applies not only to prior work, but also to our work in estimating quantities such as $\mi(\Feat;\Audio|\Text)$.} 
the differential entropies on their own are not meaningful.
By contrast, the entropies we measure, namely $\entropy(\Feat)$ and $\entropy(\Feat\mid\Channel)$ can be directly interpreted as the expected information (in bits) gained when learning the value of \Feat for a given sentence-length utterance, with or without access to the channel \Channel.

%
Second, learning probability density functions for continuous variables introduces several complications.
These prior works estimate $\entropy(\Prosody)$ using Gaussian kernel density estimator (KDE) models \citep{sheather2004density}, and
they estimate $\entropy(\Prosody\mid \Text)$ by fitting a neural network to output a parametrized distribution $\ptheta(\Prosody \mid \textvar)$ given an input text $\textvar$.
Because the KDE for $\ptheta(\Prosody)$ is nonparametric, it can fit highly complex density functions, while $\ptheta(\Prosody \mid \textvar)$ is a parametric probability density function.
This can lead to overestimates of $\entropy(\Prosody\mid\Text)$ and even negative estimates for \mi, unless specialized methods like Mixture Density Networks \citep{bishop1994mixture} are used \citep{wilcox2025using}.
On the other hand, our approach simply requires training standard neural network classifiers with cross-entropy loss.

The main advantage of the prior approach \citep{wolf2023redundancy,wilcox2025using,regev2025time} is that their models are text-only LMs and their data consists of prosodic features automatically extracted from aligned text-audio data.
Our approach requires models that accept input from the relevant channels (e.g., speech models, or multimodal text/prosody models which are currently unavailable), as well as aligned text-audio data with gold labels for the target feature \Feat.

\begin{figure*}[t]
    \centering
    \begin{subfigure}[t]{0.3\textwidth}
        \centering
        \includegraphics[width=\linewidth]{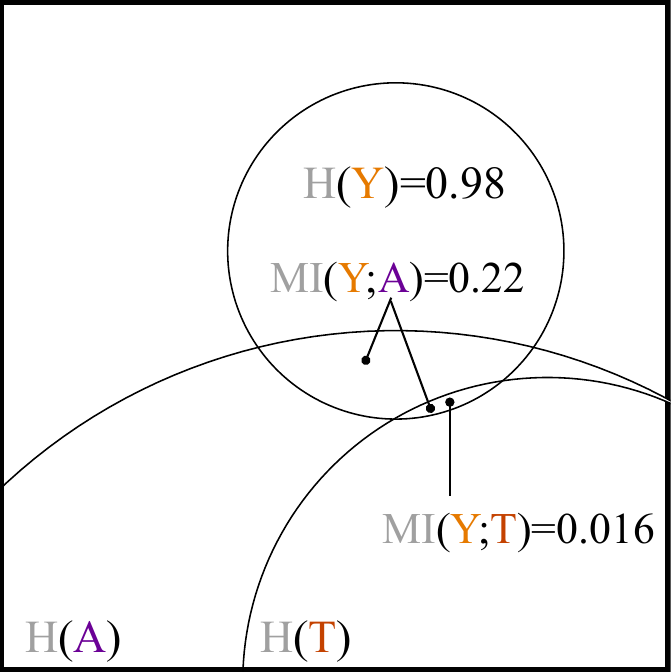}
        \caption{\Feat{=}Sarcasm}
        \label{fig:sarcasm-euler}
    \end{subfigure}
    \hfill
    \begin{subfigure}[t]{0.3\textwidth}
        \centering
        \includegraphics[width=\linewidth]{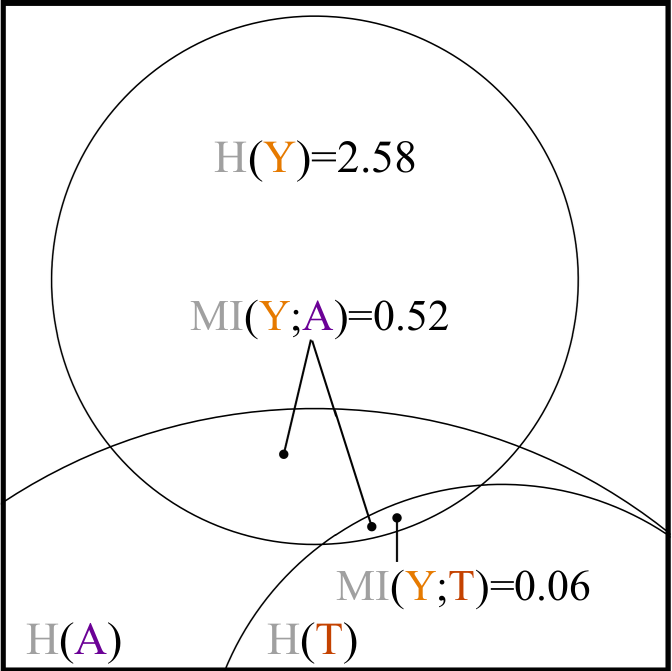}
        \caption{\Feat{=}Affect}
        \label{fig:emotion-euler}
    \end{subfigure}
    \hfill
    \begin{subfigure}[t]{0.3\textwidth}
        \centering
        \includegraphics[width=\linewidth]{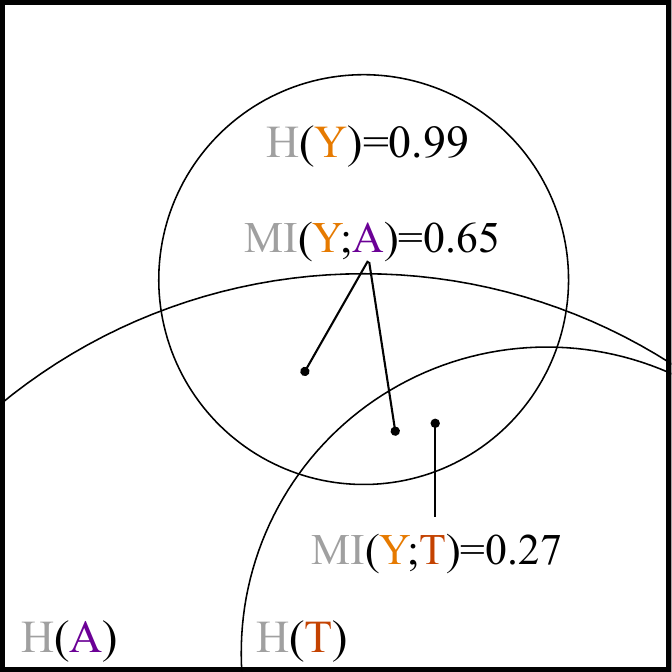}
        \caption{\Feat{=}Questionhood}
        \label{fig:question-euler}
    \end{subfigure}
    \caption{Proportional-area diagrams for sarcasm (left), affect (middle), and questionhood (right) features. In all plots, \Audio=Audio and \Text=Text. In the case of multiple arrows, the quantity refers to the sum of the areas indicated.}
    \label{fig:euler-diagrams}
\end{figure*}

\section{Case Studies}
\label{sec:meth}
We apply our mutual information estimation methods to three tasks: sarcasm, affect, and questionhood classification.
We choose the first two tasks because, besides being meaning domains that are expected to depend on prosody, they both have high-quality pre-existing datasets that pair audio, transcriptions, and labeled class data. 
For the third task, we curate a new dataset from existing resources.

\subsection{Tasks \& Datasets}


\paragraph{Sarcasm Detection}

Sarcasm, sometimes referred to as ``verbal irony,'' plays an important role in daily communication.
It can be defined as an individual expressing the opposite of what they believe \cite{haverkate1990irony}, therefore setting up incongruity between a context, an utterance, and how it is expressed \citep[i.e., its prosody;][]{MATSUI201674}.
Sarcasm serves several purposes, including modulating in-group status through humor, and its use varies between social groups \citep{dress2008sarcasm}.

One property of sarcasm that is useful for our purpose is that its relationship with text is highly ambiguous:
Some sentences are obviously sarcastic from their text (and world knowledge) alone, while others are not easy to categorize without additional prosodic information.
For example, the sentence \textit{Wow, the dog smells amazing after rolling around in the garbage!} can easily be inferred to be sarcastic.
However, the sentence \textit{I really love this place!} could be sarcastic or not, and context, gesture, or prosody can help to disambiguate this.
Such variance makes sarcasm an ideal testing bed for studying how information is distributed through different communicative channels.\looseness=-1

    

There is extensive prior work on sarcasm detection \cite{farabi2024survey}, but much of it was conducted in the text-only domain \cite{Kreuz2007LexicalIO, filatova-2012-irony, Joshi2016HarnessingSL, abercrombie-hovy-2016-putting}. 
While initial contributions have been made using multimodal models \cite{Tepperman2006yeahRS, CHEANG2008366, rakov13_interspeech}, they relied on smaller datasets.
In our study, we use the publicly available Multimodal Sarcasm Detection Dataset (MUStARD) released by \citet{castro-etal-2019-towards}.
The dataset includes 690 utterances annotated for the presence of sarcasm, drawn from a variety of situational comedy TV shows.
The classes are balanced.

\paragraph{Affect Classification}

Affect is conveyed through multiple communicative channels, including text, audio, and vision \cite{frick1985communicating, mozziconacci2002prosody, larrouy2024sound}.
Several high-quality datasets for affect detection exist \cite{Burkhardt2005ADO, Busso2008IEMOCAPIE, Bnziger2012IntroducingTG, 6681406, Russo2015THERA, Busso_2017, Lotfian_2019_3, Vidal_2020}. 
We select the MSP-Podcast dataset (version 1.12) released by \citet{Lotfian_2019_3} due to its naturalistic sources, speaker diversity, and large size.
This dataset formulates affect recognition as a classification task over 10 emotional categories. 
We refer the readers to \citet{busso2025msp} for more details regarding the categories and their distributions in the latest version of the dataset, which was released subsequent to the completion of our work. 

\paragraph{Questionhood Classification}
\label{sec:question_classification}

Questions in English can be signaled by both specific prosodic contours and by a unique syntactic form. 
In American English, yes-no questions are typically marked by rising pitch at the end, while \emph{wh}-questions are marked by falling pitch \citep{pierrehumbert1990meaning}. 
In text, questions are usually marked by subject auxiliary inversion (e.g., Are you coming?), and \emph{wh}-questions additionally contain a question word at the beginning of the sentence.
Thus, prosody and syntax convey significantly redundant information about whether an utterance is a question.
An exception is echo questions, such as \emph{You're leaving already?}, where the prosody is the only cue to questionhood (after removing punctuation).
Consequently, we predict that prosody will add comparatively little information in addition to text when determining the questionhood of an utterance.

We formulate questionhood classification as a binary classification task, classifying utterances as questions or non-questions (assertions, imperatives, etc.).%
\footnote{This task is related to dialogue act classification \citep[e.g.,][]{stolcke2000dialogue}, but we adopt a vastly simplified set of classes to interpret the results easily.}
The data is sampled from the MSP-Podcast dataset \citep{Lotfian_2019_3}.
We segment the transcripts by sentence using spaCy \citep{honnibal2020spacy}, align the audio using the Montreal Forced Aligner \citep{mcauliffe2017montreal}, and remove all utterances that are shorter than 2 seconds.
Any utterance with a transcript ending in a question mark is labeled as a question, and all remaining utterances are labeled as non-questions. 
Finally, we downsample non-questions so that the dataset contains an approximately equal number of questions and non-questions.\footnote{We also downsample non-questions in such a way that the distribution of durations remains the same as that of the questions.} The train, development, and test splits have 13842, 1538, and 3845 samples, respectively.
All occurrences of [.,?] are removed prior to training.

\begin{figure*}[t]
    \centering
    \begin{minipage}[t]{0.31\textwidth}
        \begin{subfigure}{\textwidth}
            \centering
            \includegraphics[width=\textwidth]{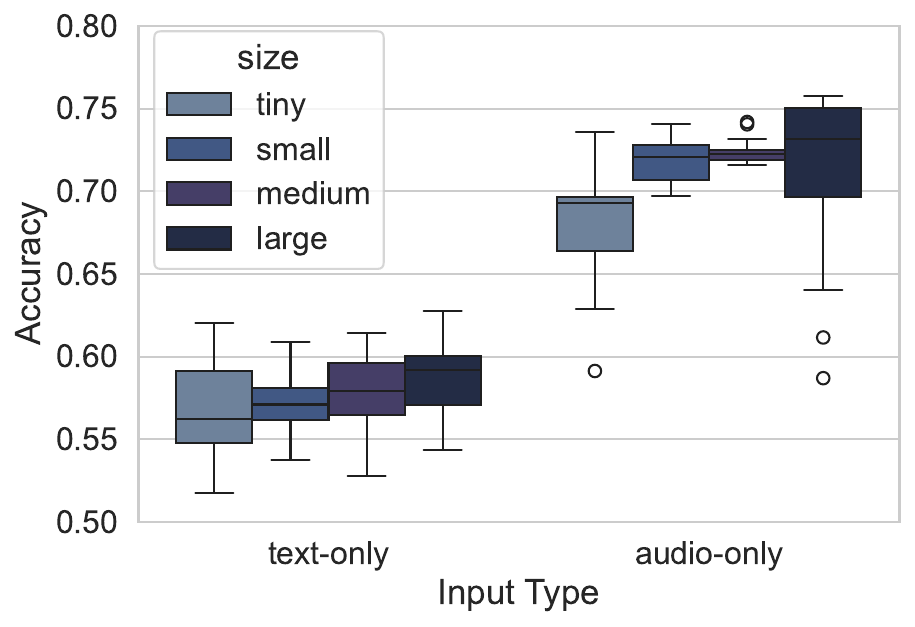}
            \caption{Test acc.~on sarcasm detection.}
            \label{fig:sarcasm-acc}
        \end{subfigure}
                
        \begin{subfigure}{\textwidth}
            \centering
            \includegraphics[width=\textwidth]{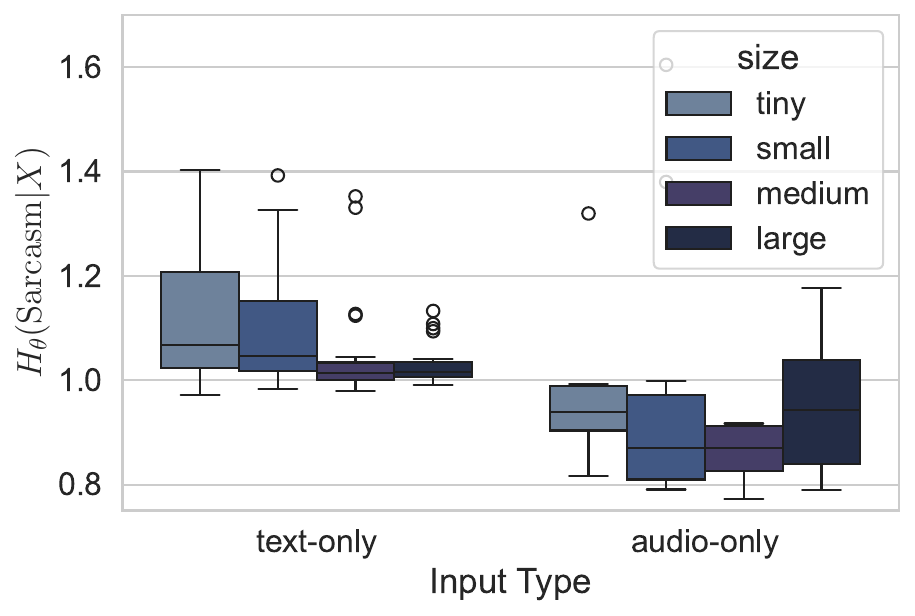}
            \caption{Test loss on sarcasm detection.}
            \label{fig:sarcasm-loss}
        \end{subfigure}
    \end{minipage}
    \hfill
    \begin{minipage}[t]{0.31\textwidth}
        \begin{subfigure}{\textwidth}
            \centering
            \includegraphics[width=\textwidth]{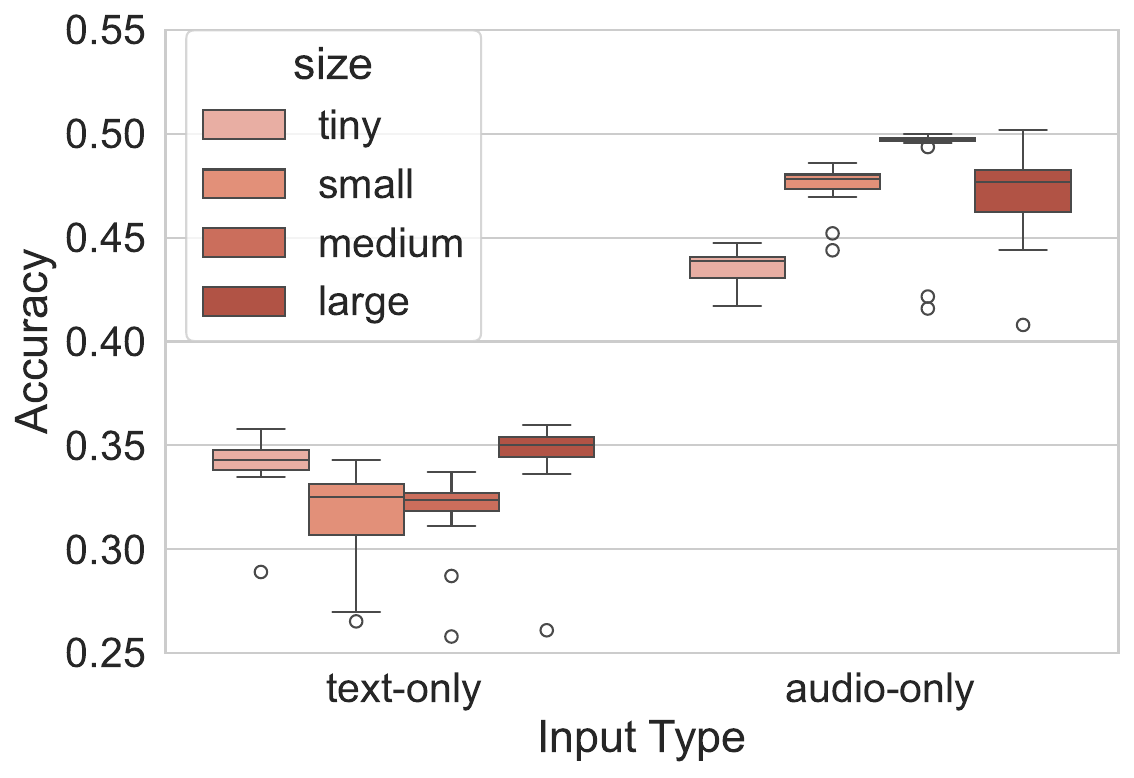}
            \caption{Test acc.~on affect classification.}
            \label{fig:emotion-acc}
        \end{subfigure}
        \begin{subfigure}{\textwidth}
            \centering
            \includegraphics[width=\textwidth]{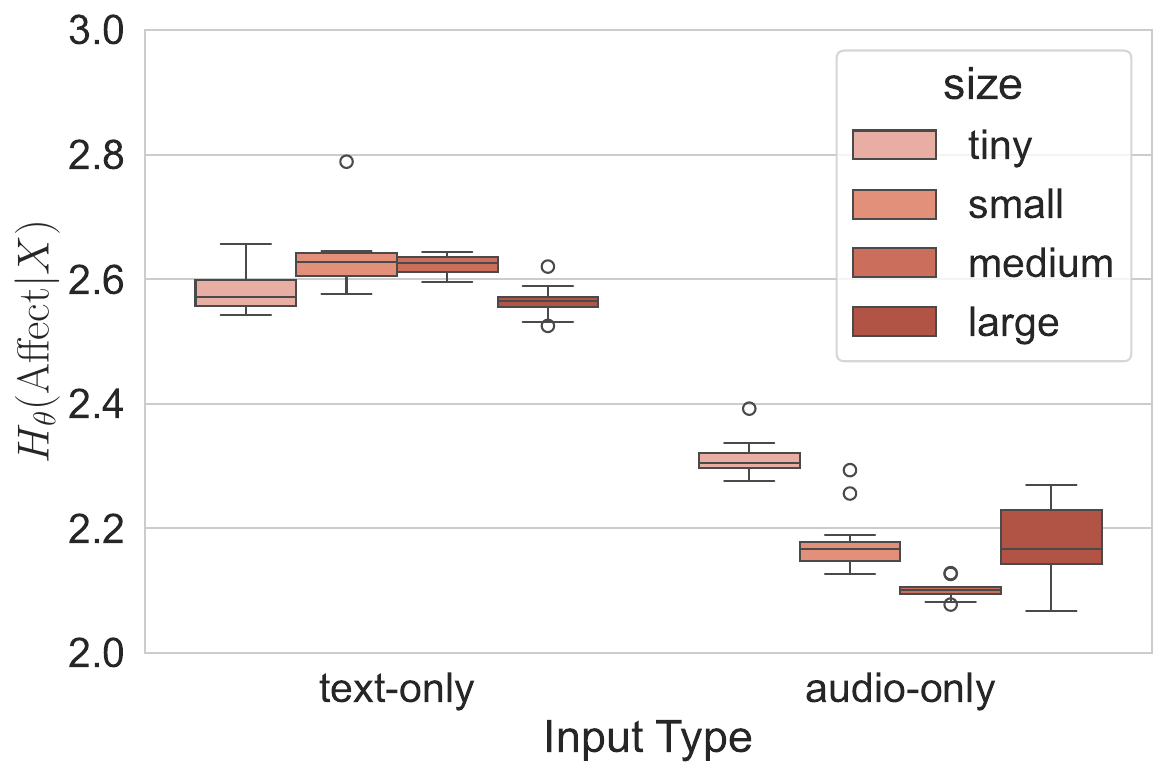}
            \caption{Test loss on affect classification.}
            \label{fig:emotion-loss}
        \end{subfigure}
    \end{minipage}
    \hfill
    \begin{minipage}[t]{0.31\textwidth}
        \begin{subfigure}{\textwidth}
            \centering
            \includegraphics[width=\textwidth]{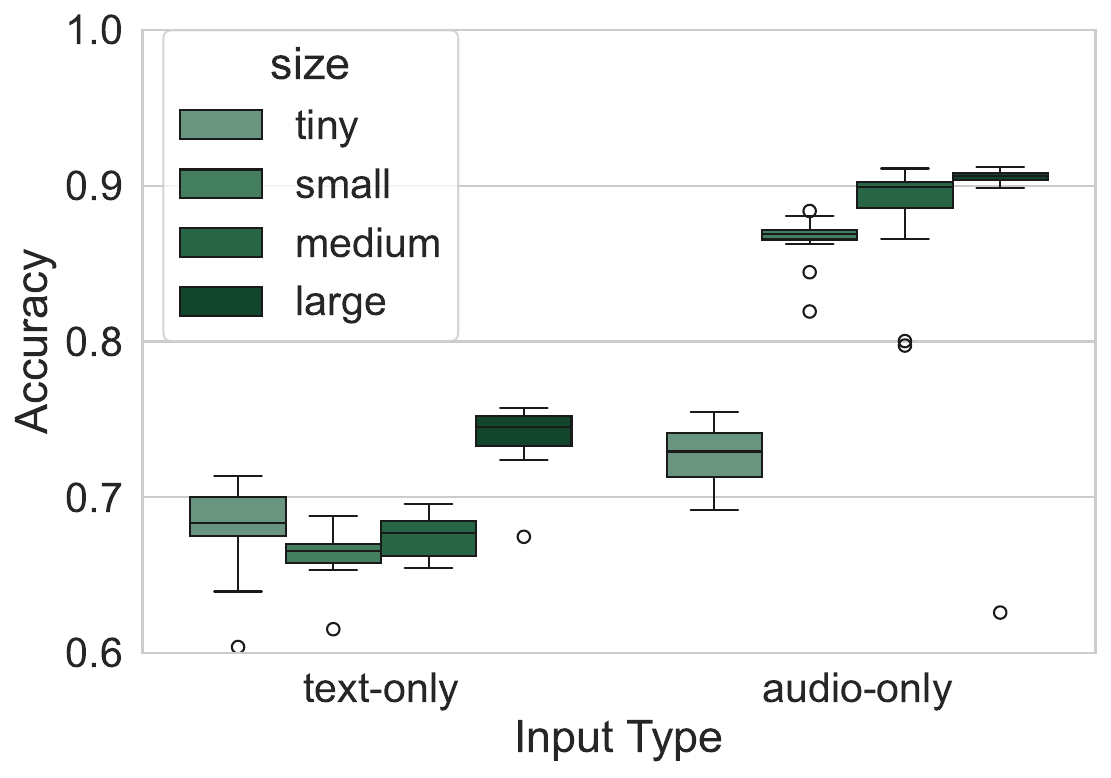}
            \caption{Test acc.~on question classification.}
            \label{fig:question-acc}
        \end{subfigure}
        \begin{subfigure}{\textwidth}
            \centering
            \includegraphics[width=\textwidth]{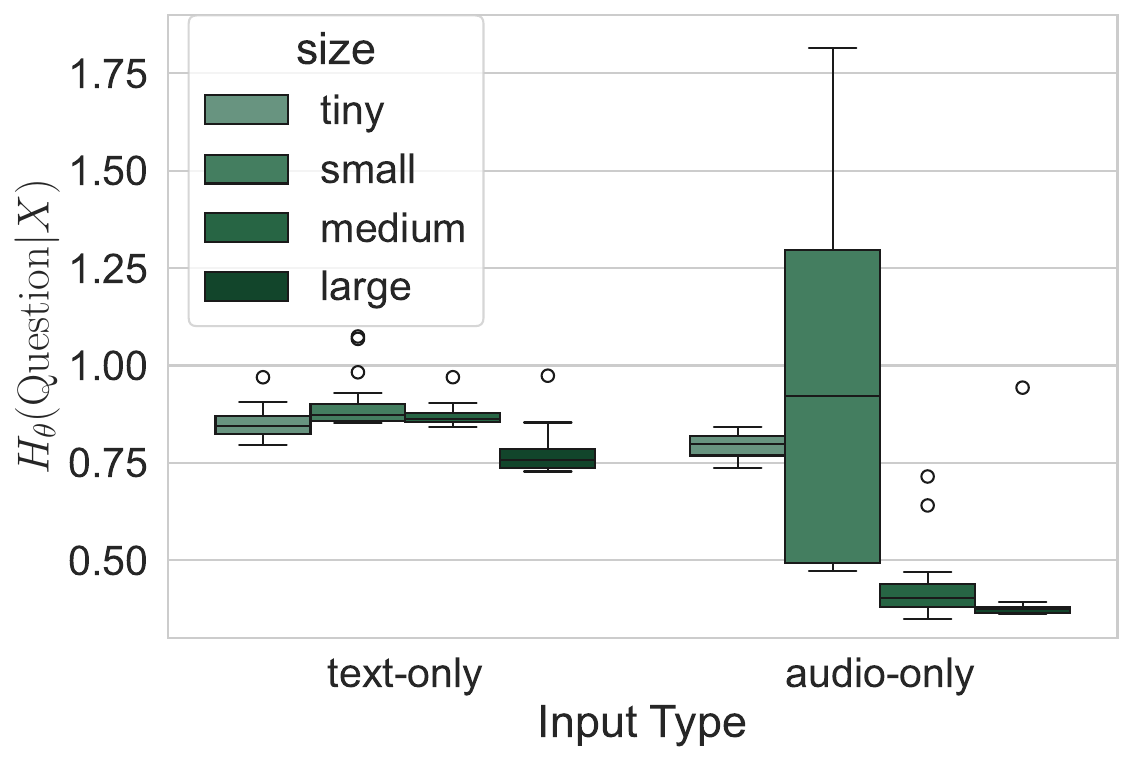}
            \caption{Test loss on question classification.}
            \label{fig:question-loss}
        \end{subfigure}
    \end{minipage}
    
    \caption{Test set performance on sarcasm (left) and affect (middle), and questionhood (right) classification tasks.}
    \label{fig:acc-loss-comparison}
\end{figure*}

\subsection{Experimental Setup}

As described in \Cref{sec:estimation_methods}, we estimate conditional entropy $\entropy(\Feat\mid\Channel)$ as the cross-entropy of a classifier trained to predict \Feat given \Channel.
We train classifiers by fine-tuning pretrained Transformer models.
We base our code on the Transformers library pipeline \citet{wolf2019huggingface}.
We experiment with four different model sizes for each model family and update all model parameters during training.
We fine-tune the model using the cross-entropy loss and Adam optimizer \cite{kingma2014adam}.
We train for a fixed number of epochs and select the checkpoint that has the lowest loss on our development split.
To select hyperparameters, we perform a sweep of 20 runs for each model size and select the top model based on the test loss value.
For sarcasm detection, we implement 5-fold cross-validation due to the dataset's small size. 
For affect and questionhood classification, we use train, development, and test splits.
More information on our hyperparameters is provided in Appendix \ref{sec:appendix_hp}.

\paragraph{Text-only models} For the text-only models, we fine-tune the GPT-2 suite of models \cite{Radford2019LanguageMA} on the transcript portion of the data for each task.
We use small, medium, large, and xl models.\footnote{For consistency in naming conventions across modalities, we subsequently refer to these models as tiny, small, medium, and large, respectively.}
We train a classification head on top of this model.
For sarcasm detection we fine-tune the entire model, and for affect and questionhood classification, we use low-rank adaptation (LoRA) \cite{hu2022lora} to reduce training time.


\paragraph{Audio-only models} For our audio-only models, we use Whisper \cite{radford2022whisper} and wav2vec 2.0 \citep{baevski2020wav2vec}.
Whisper is a state-of-the-art multilingual speech-to-text model based on a Transformer \cite{NIPS2017_3f5ee243} encoder-decoder architecture.
The encoder takes the $\log$ mel spectrogram as input and the decoder uses text tokens as input and output.
Since our tasks involve classifying an audio recording, as opposed to generating text, we fine-tune only the \emph{encoder} portion to get the audio representations to pass onto a classification layer.
We use Whisper tiny, small, medium, and large sized models. 
We also experiment with wav2vec \cite{baevski2020wav2vec} for fine-tuning on audio tasks, as it is pretrained on an audio-only masking objective, unlike Whisper, which is trained mainly to perform speech recognition on paired text-audio data.
Finally, although we assume that the audio \Audio contains all the information about the text \Text, in case audio models in practice fail to encode some information about \Text, we conduct additional experiments fine-tuning \texttt{audio+text} models accepting both audio and text input.
These models are described in Appendix \ref{sec:audio_text}.


\section{Results}

\paragraph{Information Quantification} The information diagrams for sarcasm, affect, and questionhood classification are shown in \cref{fig:euler-diagrams}.
They show proportional areas for $\entropy(\Feat)$, $\mi(\Feat;\Audio)$, and $\mi(\Feat;\Text)$.
The regions $\entropy(\Audio)$ and $\entropy(\Text)$ are not estimated and are purely illustrative.

For both sarcasm and affect, we observe that the mutual information between the feature of interest and the audio $\mi(\Feat;\Audio)$ is larger than its mutual information with the text only $\mi(\Feat;\Text)$ by over an order of magnitude. 
This suggests that the audio contains substantial information about both of these features above and beyond the text, both confirming our intuitions and corroborating previous findings \citep{Tepperman2006yeahRS, CHEANG2008366}.
While $\mi(\Feat;\Audio)$ values for sarcasm and affect are quite different, we note that the uncertainty coefficients $\frac{\mi(\Feat;\Audio)}{\entropy(\Feat)}$ \citep{press2007numerical} are quite similar -- 0.22 and 0.20 respectively -- telling us that the audio signal conveys a similar proportion of the total information in these features.

By contrast, for questionhood, $\mi(\Feat;\Audio)$ is only about 2.4 times greater than $\mi(\Feat;\Text)$.
This aligns with our prediction that prosody would provide relatively less additional information compared to other tasks, reflecting the fact that questions are often indicated by syntax, in addition to prosody.

We also estimate the information that prosody \textit{uniquely} conveys about sarcasm, affect, and questionhood, $\mi(\Feat;\Prosody\mid\Text)$, as discussed in Section \ref{sec:approach}.
Subtracting the mutual information of the features with the text from its mutual information with the audio provides this estimate.
This quantity is 0.20 and 0.46 bits for sarcasm and affect, respectively, making up a large fraction of the $\mi(\Feat;\Audio)$.
It is 0.38 bits for questionhood, making up a smaller portion of the $\mi(\Feat;\Audio)$.

We observe a large difference in $\entropy(\Feat)$ for sarcasm and questionhood on the one hand and affect on the other, where for affect it is larger. 
While this matches intuitions -- affect is inherently higher-dimensional than sarcasm or questionhood -- this is also a consequence of the labeling scheme:
The sarcasm and questionhood datasets have binary classes, while the affect dataset has ten classes.\looseness=-1

\paragraph{Classifier Performance}

We find several patterns of interest in the performance of the classifiers trained to make the above estimates.
Since cross-entropy provides an upper bound on entropy, i.e., $\xent(\Feat\mid\Channel) \geq \entropy(\Feat\mid\Channel)$, we use the lowest-loss runs to estimate mutual information values in \cref{fig:euler-diagrams}. 
For $\mi(\Feat;\Text)$, we select the best \texttt{text-only} model run across all model sizes, and for $\mi(\Feat;\Audio)$, we use the lowest loss run among the \texttt{audio-only} models and sizes.\todo[replace `\texttt{audio-only}` with macro.]

We report the distribution of losses and accuracies for GPT-2 and Whisper models over 20 runs in \cref{fig:acc-loss-comparison}. 
To calculate the accuracy for one sarcasm run, we take the average accuracy across all 5-folds for each run. 
In the Appendix, we report additional results for wav2vec 2.0 (\cref{fig:audio-only-acc-loss-comparison}) and for \texttt{audio+text} models (\cref{fig:audio-text-acc-loss-comparison}).
First, we find that our classifiers have strong performance relative to baselines:
Our best sarcasm classifier shows a 10-point improvement over the baseline released by \citet{castro-etal-2019-towards}, and our best emotion classifier is competitive with other models reported by \citet[][Table 5]{goncalves2024odyssey}.\looseness=-1

Second, we observe that the figures show several similar trends: 
Larger \texttt{audio-only} models tend to have stronger performance (though Whisper large does not consistently outperform Whisper medium), while model size has a limited impact on the performance of \texttt{text-only} models, though exceptions exist.
In all cases, \texttt{audio-only} models have higher accuracy than the \texttt{text-only} models. 
In \cref{fig:audio-only-acc-loss-comparison} and  \cref{fig:audio-text-acc-loss-comparison}, we see that \texttt{audio-only} models based on Whisper medium or -large have stronger performance than wav2vec 2.0 and \texttt{audio+text} models.


\section{Discussion and Conclusion}

The approach we introduce in Section \ref{sec:approach} enables us to characterize how a particular kind of information is distributed over multiple communication channels. 
As case studies, we investigated how information about sarcasm, affect, and questionhood are conveyed through prosody and text.
Our results provide nuance on the importance of prosody for these tasks: 
We find that the amount of information conveyed by the audio signal can be over an order of magnitude larger than the information conveyed by text alone, but this is highly dependent on the feature of interest. 
This result highlights the importance of treating language as a multi-channel communication system, and suggests that important properties of language will be overlooked if it is analyzed only in one channel (i.e., text) as is common.\looseness=-1

An important caveat is that we restrict both audio and text to a single sentence.
If we define $\Text'$ and $\Prosody'$ to be random variables over the text and prosody for the rest of the discourse, it stands to reason that $\mi(\Feat;\Text,\Text')$ would be significantly larger than $\mi(\Feat;\Text)$ for both sarcasm and affect, i.e., the larger context adds significantly more information about the feature of interest than one can get from the sentence alone. 
Indeed, \citet{castro-etal-2019-towards} show that previous utterance information improves sarcasm detection.
On the other hand, $\mi(\Feat;\Prosody,\Prosody')$ might be similar to $\mi(\Feat;\Prosody)$, as prior work suggests that the unique information conveyed by prosody, relative to its past text, is mostly local \citep{regev2025time}.
Thus, our results show that while \emph{local} textual information is often insufficient to judge sarcasm or affect, local \emph{prosody} often is sufficient.
This conclusion has implications for discourses lacking substantial textual context: For example, long-term context may not be available for a person reading a short text message or for an automated speech-to-text customer service system responding to a customer over the phone.
A fine-grained comparison of the information contributions of text and prosody in short and long contexts is a promising direction for future work.\looseness=-1

This work's main contribution is to introduce a general framework, which future work can extend to other domains of meaning (\Feat), communicative channels, and languages.
Regarding other domains of meaning, there is an extensive prior literature on the use of prosody for syntactic disambiguation \citep{selkirk2011syntax,pauker2011prosody} and word boundary detection \citep{cutler2014prosody}, with important implications for language acquisition \citep{soderstrom2003prosodic}.
Prosody also provides important cues about turn-taking \citep{ruiter2006projecting,gravano2011turn,cutler2018analysis} and back-channels \citep{clark2025relationship}.

Other communicative channels that one might extend the approach to are specific prosodic features and combinations thereof. 
This will require architectural and representational innovations, as it will be necessary to train classifiers that take prosodic representations as input.
Using vision language models \citep[e.g.,][]{radford2021learning}, it will also be possible to investigate information conveyed through \emph{visual} channels, such as facial expressions and hand gestures.

Finally, whereas we limit the present study to English, future work should compare a more diverse set of languages.
Recent work \citep{wilcox2025using} has suggested that prosodic typologies can be investigated from an information-theoretic lens. 
Our approach can provide a finer-grained technique to identify how typologically distinct languages use the prosodic channel to convey different aspects of linguistic information.


\section{Limitations}\label{sec:limitations}

In  \Cref{sec:approach}, we argue that \legendCircleNine $\mi(\Feat;\Audio\mid\Text)$ serves as a reasonable approximation of \legendCircleTwo $\mi(\Feat;\Prosody\mid\Text)$, since the only information captured by \legendCircleNine but not by \legendCircleTwo reflects mostly irrelevant factors such as speaker identity or background noise. 
However, to more precisely estimate the contribution of the prosodic channel to a particular linguistic phenomenon -- and, by extension, to human communication -- we would ideally measure $\mi(\Feat;\Prosody\mid\Text)$ directly. 
This would also allow us to estimate two additional quantities of interest: (1) the unique contribution of text to \Feat, $\mi(\Feat;\Text\mid\Prosody)$, and (2) the redundant contributions of text and prosody to \Feat, $\mi(\Feat;\Text;\Prosody)$.
However, current spoken language models do not accept only prosodic features as input, making this estimation impossible. 
We leave it to future work to tackle the substantial problem of designing a new architecture that can take representations of isolated prosodic features as input,
enabling us to measure all of these quantities directly, either for individual prosodic features such as pitch, or the entire prosodic channel.

It is important to acknowledge that our feature estimates are heavily influenced by idiosyncrasies of the languages, datasets, and models we study. 
As we study these tasks in English only, our specific quantitative findings may not generalize to typologically different languages such as tone or pitch accent languages.
For sarcasm detection, we rely on a balanced dataset in which sarcastic and non-sarcastic utterances are equally represented. 
However, this setup does not mimic our real-world human conversations -- sarcastic utterances are relatively rare. 
In the case of affect recognition, we work in a highly constrained setting where each utterance is labeled with a single emotion. 
However, a single utterance may convey multiple emotions simultaneously or to varying degrees. 
Moreover, the dataset includes only ten discrete emotions -- a small subset of the full range emotions that humans can express.
While these specifics may arguably have relatively little impact on our estimates of ratio of $\mi(\Feat;\Audio)$ and $\mi(\Feat;\Text)$, they strongly affect the magnitude of our estimates, as more imbalanced labels will lead to lower unconditional entropy, and more fine-grained labeling schemes will lead to higher unconditional entropy.

Finally, we acknowledge that we do not fine-tune the most powerful text LLMs available. 
Fine-tuning models larger than GPT-2 XL could lead to better estimations of $\mi(\Feat; \Text)$.
However, we opt not to work with these models due to the computational demands of fine-tuning.
While it is possible that Whisper large would achieve state-of-the-art performance on these tasks, there is rapid progress in spoken language modeling, meaning better estimates may be achievable in the future.
We leave it to future work to more thoroughly investigate the impact of the specific model choice on \mi estimation.\looseness=-1


\section*{Acknowledgments}
We thank Sarenne Wallbridge for our early conversations about this paper. This work used resources available through the National Research Platform (NRP) at the University of California, San Diego \citep{nautilus}. NRP has been developed, and is supported in part, by funding from National Science Foundation, from awards 1730158, 1540112, 1541349, 1826967, 2112167, 2100237, and 2120019, as well as additional funding from community partners.

\bibliography{custom}

\newpage
\newpage

\appendix
\newpage

\vphantom{.}
\vspace{90em}
\section{Hyperparameters}
\label{sec:appendix_hp}

We use Bayesian hyperparameter optimization to run our sweeps, optimizing to minimize the test loss. For all experiments, we search for the optimal learning rate using a log-uniform distribution. In the rest of this section, we share the task-specific hyperparameter choices we have made and the compute used. 

\paragraph{Affect Classification} Table \ref{tab:affect_detection} shows the hyperparameters used for this task. All experiments use cosine learning rate decay for smooth convergence and better generalization. Our preliminary experiments on \texttt{audio-text} models show that our choice of batch size and number of epochs does not affect task performance. Therefore, we choose values that optimize training time.

We use V100 GPUs and A100 GPUs to perform our \texttt{text-only} and \texttt{audio-only} affect recognition sweeps respectively. To conduct our \texttt{audio-text} tiny, small, and medium affect recognition model sweeps we use 8 A10, 4 A6000, and 2 A100 GPUs respectively. Our \texttt{text-only} model sweeps took less than 2 days each. Our \texttt{audio-only} sweeps took 3 to 7 days depending on the model size. Finally, our \texttt{audio-text} sweeps took 5 to 10 days depending on the model size.


\begin{table}[h!]
\resizebox{\columnwidth}{!}{%
\begin{tabular}{llll}
    \toprule
    \textbf{Hyperparameters} & \textbf{text-only} & \textbf{audio-only} & \textbf{audio-text} \\
    \midrule
    {Epochs} & 5, 10, 15 & 5, 10, 15 & 5 \\
    {Batch Size} & 8, 16, 32, 64 & 8, 16, 32, 64 & 64 \\ 
    {Weight Decay} & 0, 0.01, 0.1 & 0, 0.01, 0.1 & 0, 0.01, 0.1 \\
    {Learning Rate} & [1e-6, 1e-4] & [1e-6, 1e-4] & [1e-6, 1e-4] \\
    {LoRA $r$} & 4, 8, 16 & -- & -- \\
    {LoRA $\alpha$} & 8, 16, 32 & -- & -- \\
    \bottomrule
    \end{tabular}
}
\caption{Affect Classification Hyperparameters}
\label{tab:affect_detection}
\end{table}

\paragraph{Sarcasm Detection} Table \ref{tab:sarcasm_detection} shows the hyperparameters used for this task. We use A100 GPUs to train all our sarcasm detection models. No sweep across all modalities and sizes ran for more than 2 days.

\begin{table}[h!]
\resizebox{0.6\columnwidth}{!}{%
    \begin{tabular}{ll}
    \toprule
    \textbf{Hyperparameters} & \textbf{All Models} \\
    \midrule
    Epochs & 5, 10, 15  \\
    Batch Size & 8, 16, 32, 64  \\ 
    Weight Decay & 0, 0.01, 0.1  \\
    Learning Rate & [1e-6, 1e-4] \\
    \bottomrule
    \end{tabular}
}
\caption{Sarcasm Detection Hyperparameters}
\label{tab:sarcasm_detection}
\end{table}

\paragraph{Questionhood Classification}  With the exception of Whisper large, Table \ref{tab:question_classification} shows the hyperparameters used for all the models for this task. For Whisper large, to maximize the usage of the available GPU VRAM and thereby reduce the training time, we run our sweep on the following (effective) batch sizes: 16, 32, 64, 128. Our preliminary experiments on Whisper large show that our choice ofnumber of epochs does not affect task performance. Therefore, we limit the number of epochs for this model to be 5.
The rest of the hyperparameters are the same as the other models. We use A100 and V100 GPUs to train all our questionhood classification models. No sweep across all modalities and sizes ran for more than 3 days.

\begin{table}[h!]
\resizebox{\columnwidth}{!}{%
    \begin{tabular}{lll}
    \toprule
    \textbf{Hyperparameters} & \textbf{text-only} & \textbf{audio-only}  \\
    \midrule
    {Epochs} & 5, 10, 15 & 5, 10, 15 \\
    {Batch Size} & 8, 16, 32, 64 & 8, 16, 32, 64 \\
    {Weight Decay} & 0, 0.01, 0.1 & 0, 0.01, 0.1 \\
    {Learning Rate} & [1e-6, 1e-4] & [1e-6, 1e-4] \\
    {LoRA $r$} & 4, 8, 16 & --  \\
    {LoRA $\alpha$} & 8, 16, 32 & --  \\
    \bottomrule
    \end{tabular}
}
\caption{Questionhood Classification Hyperparameters}
\label{tab:question_classification}
\end{table}

\section{Audio+Text models}\label{sec:audio_text}
We conduct additional experiments with models that accept both audio and text input for sarcasm and affect classification. 
The motivation is that, while the audio \Audio theoretically contains all the information about the text \Text, audio models may fail to encode some of this information.
If this is the case, we can mitigate this problem by inputting the full text in addition to the audio.

For our \texttt{audio+text} models, we use representations from the encoder part of the Whisper model to get audio embeddings and representations from the GPT-2 model to get the text embeddings. 
We concatenate both these embeddings and use the combination as input to a classification layer.
We choose this architecture so that the results would be easily comparable to the \texttt{audio-only} and \texttt{text-only} models. 
We pair GPT-2 small and Whisper tiny, GPT-2 medium and Whisper small, and GPT-2 large and Whisper medium. 
We refer to these pairs as \texttt{audio+text} tiny, small, and medium respectively.

\begin{figure*}[t]
    \centering
    \begin{minipage}[t]{0.45\textwidth}
        \begin{subfigure}{\textwidth}
            \centering
            \includegraphics[width=\textwidth]{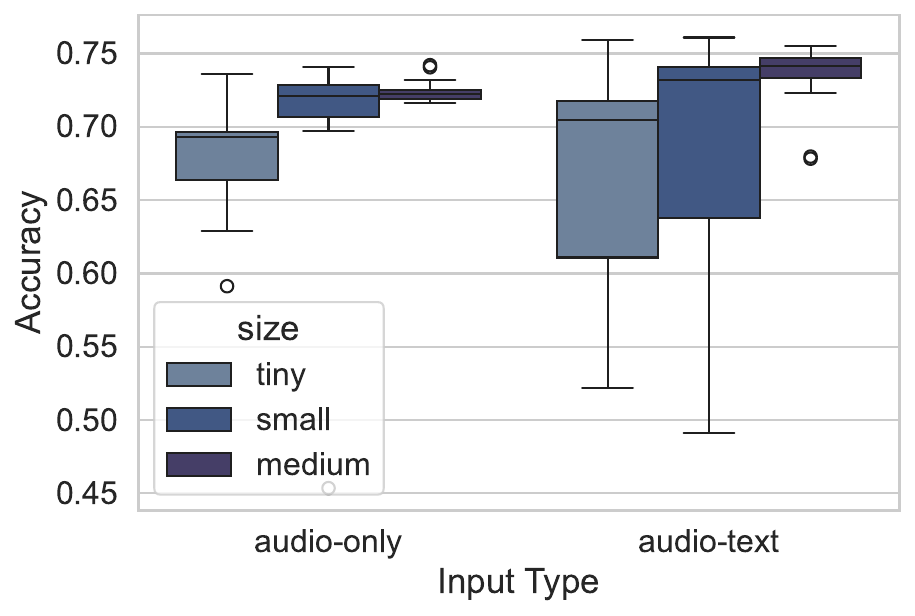}
            \caption{Test acc.~on sarcasm detection.}
            \label{fig:audio-text-sarcasm-acc}
        \end{subfigure}
                
        \begin{subfigure}{\textwidth}
            \centering
            \includegraphics[width=\textwidth]{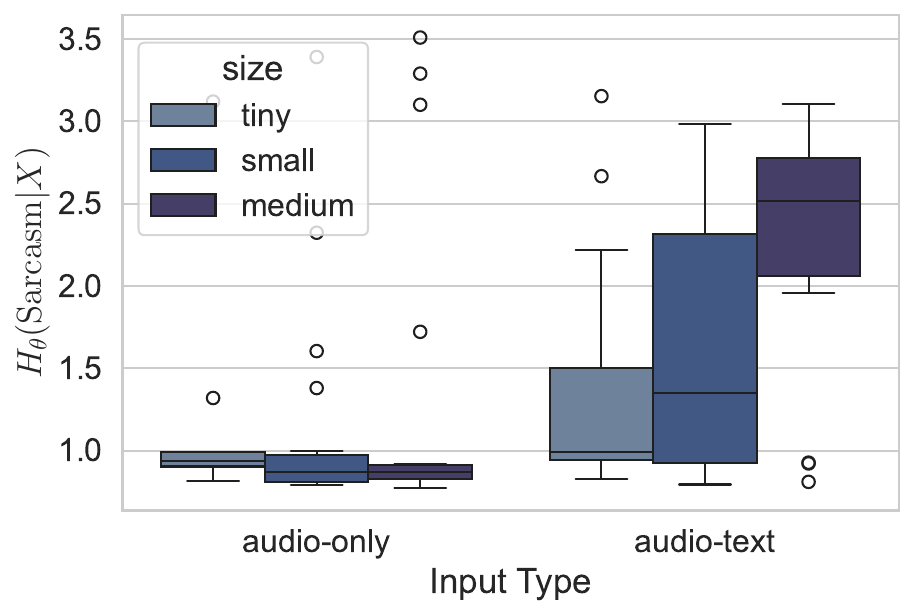} 
            \caption{Test loss on sarcasm detection.}
            \label{fig:audio-text-sarcasm-loss}
        \end{subfigure}
    \end{minipage}
    \hfill
    \begin{minipage}[t]{0.45\textwidth}
        \begin{subfigure}{\textwidth}
            \centering
            \includegraphics[width=\textwidth]{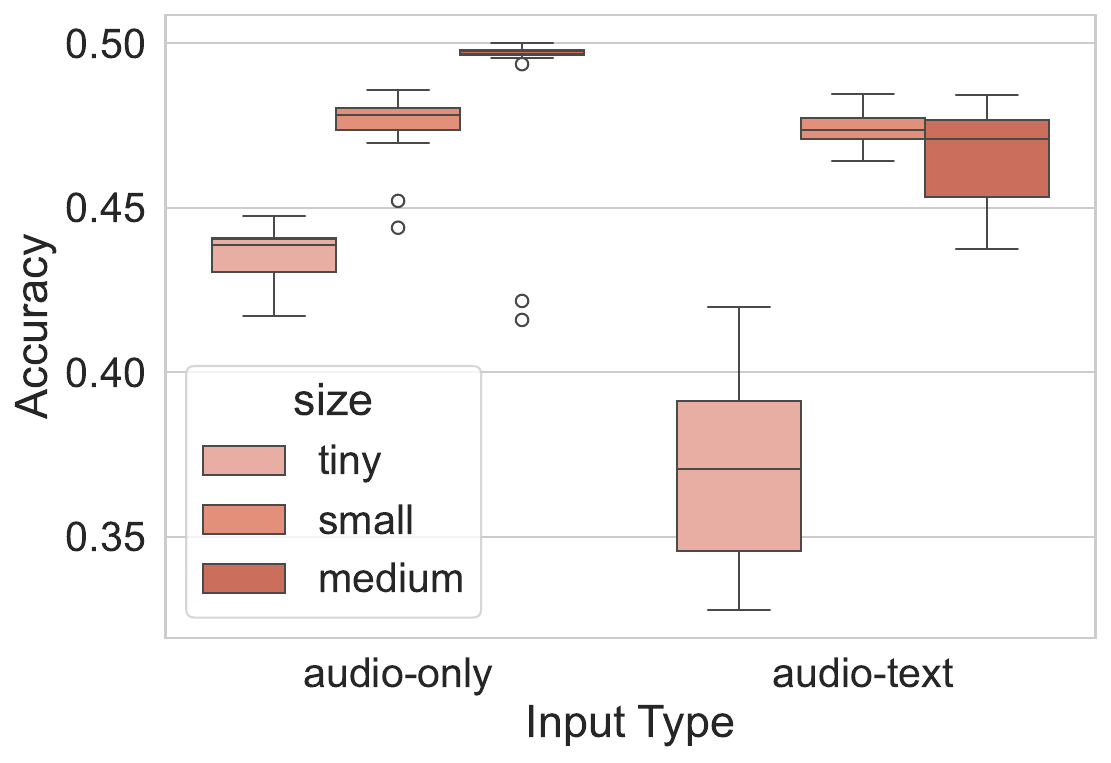}
            \caption{Test acc.~on affect classification.}
            \label{fig:audio-text-emotion-acc}
        \end{subfigure}
        \begin{subfigure}{\textwidth}
            \centering
            \includegraphics[width=\textwidth]{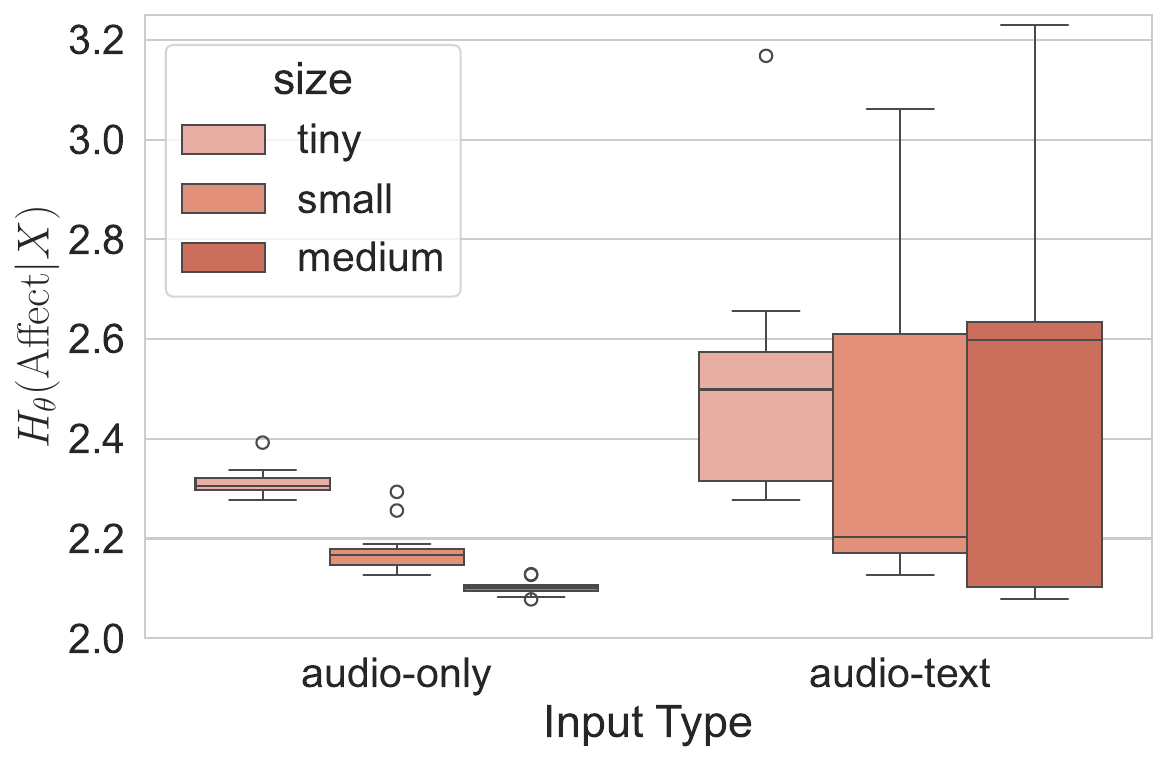}
            \caption{Test loss on affect classification.}
            \label{fig:audio-text-emotion-loss}
        \end{subfigure}
    \end{minipage}
    \caption{\texttt{audio-only} and \texttt{audio-text} performance on sarcasm (left) and affect (right)}
    \label{fig:audio-text-acc-loss-comparison}
\end{figure*}

In \cref{fig:audio-text-emotion-loss}, we present losses and accuracies for models.




\begin{figure*}[t]
    \centering
    \begin{minipage}[t]{0.31\textwidth}
        \begin{subfigure}{\textwidth}
            \centering
            \includegraphics[width=\textwidth]{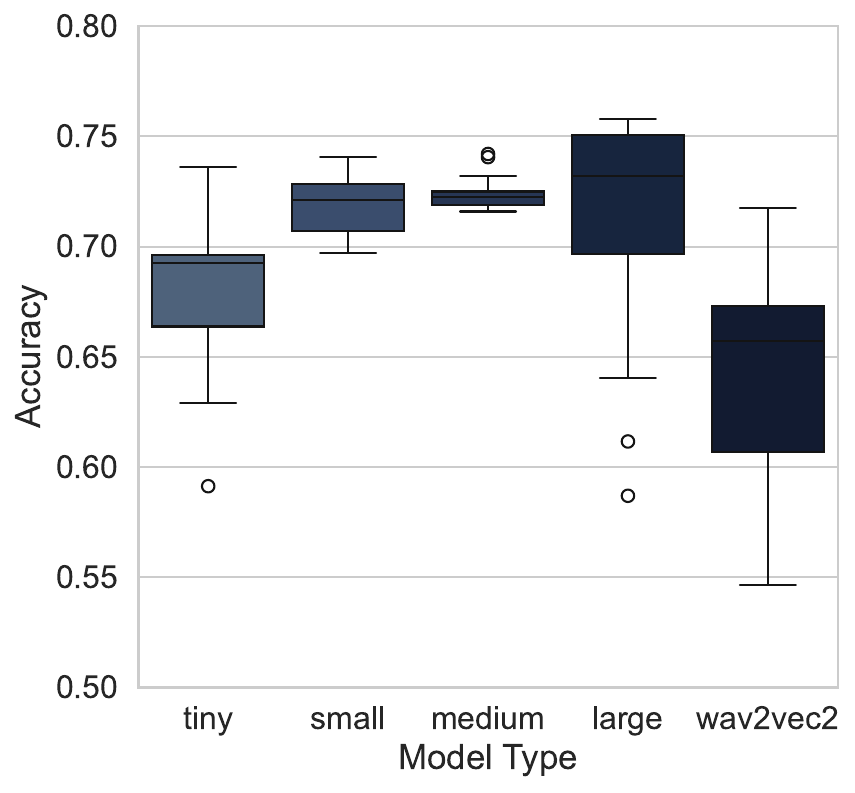}
            \caption{Test acc.~on sarcasm detection.}
            \label{fig:audio-only-sarcasm-acc}
        \end{subfigure}
                
        \begin{subfigure}{\textwidth}
            \centering
            \includegraphics[width=\textwidth]{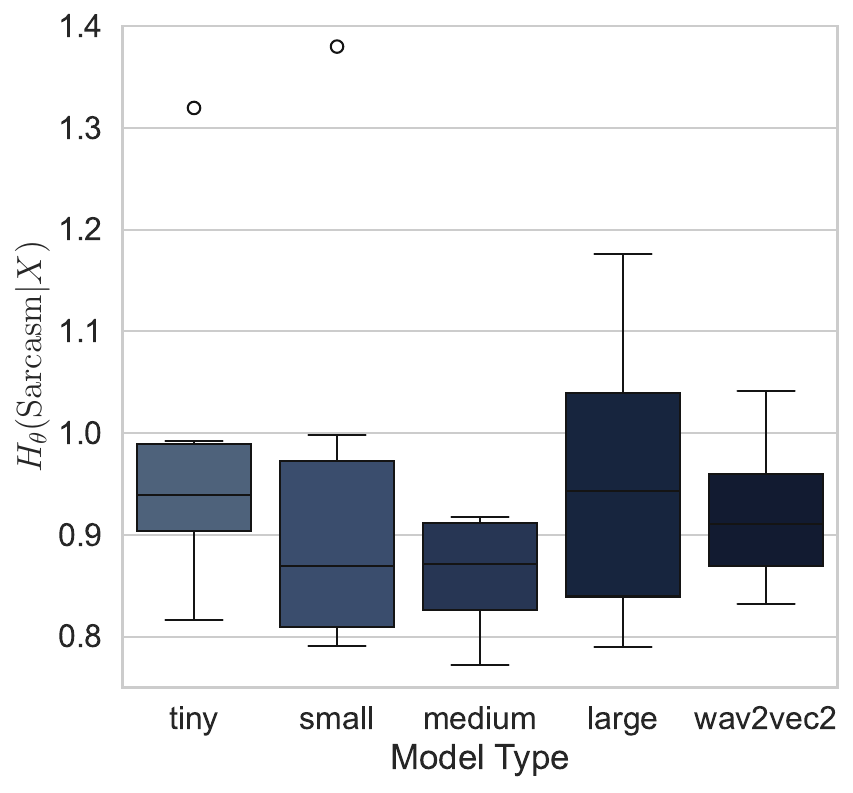} 
            \caption{Test loss on sarcasm detection.}
            \label{fig:audio-only-sarcasm-loss}
        \end{subfigure}
    \end{minipage}
    \hfill
    \begin{minipage}[t]{0.31\textwidth}
        \begin{subfigure}{\textwidth}
            \centering
            \includegraphics[width=\textwidth]{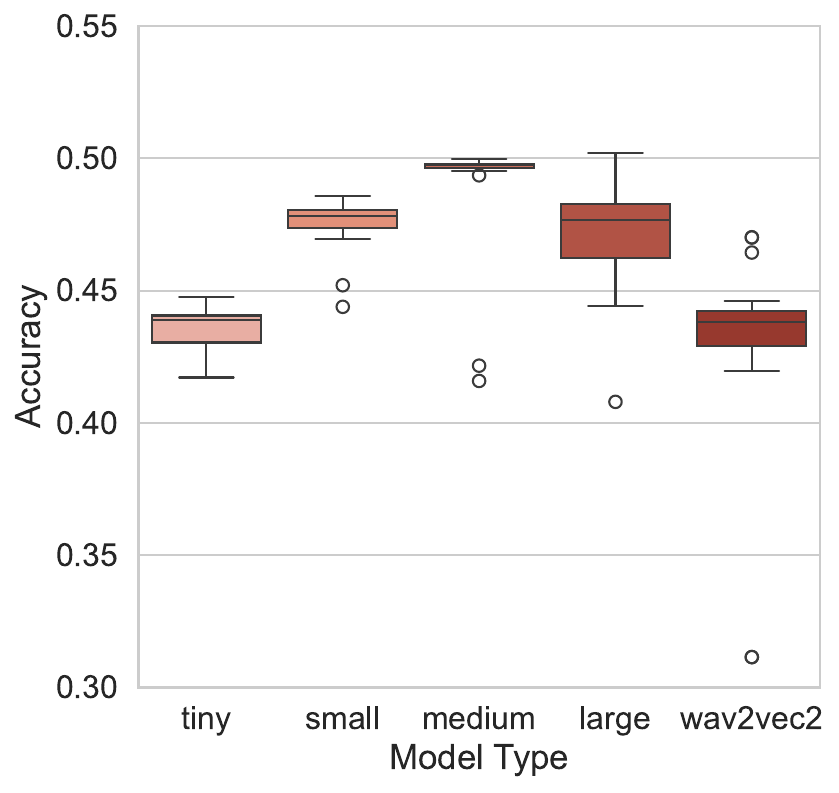}
            \caption{Test acc.~on affect classification.}
            \label{fig:audio-only-emotion-acc}
        \end{subfigure}
        \begin{subfigure}{\textwidth}
            \centering
            \includegraphics[width=\textwidth]{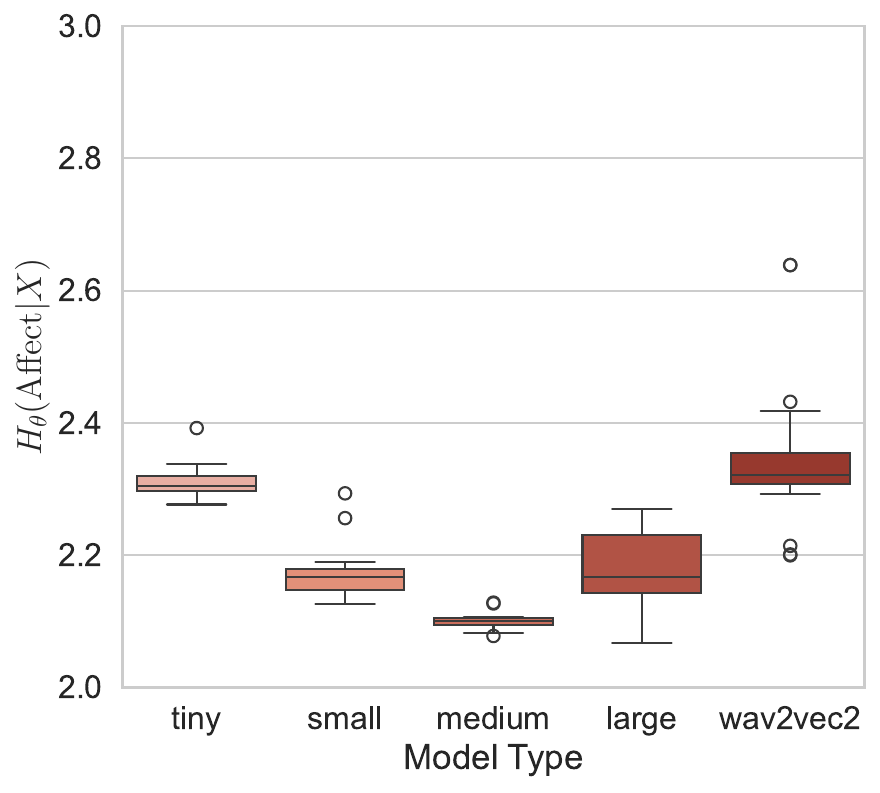}
            \caption{Test loss on affect classification.}
            \label{fig:audio-only-emotion-loss}
        \end{subfigure}
    \end{minipage}
    \hfill
    \begin{minipage}[t]{0.31\textwidth}
        \begin{subfigure}{\textwidth}
            \centering
            \includegraphics[width=\textwidth]{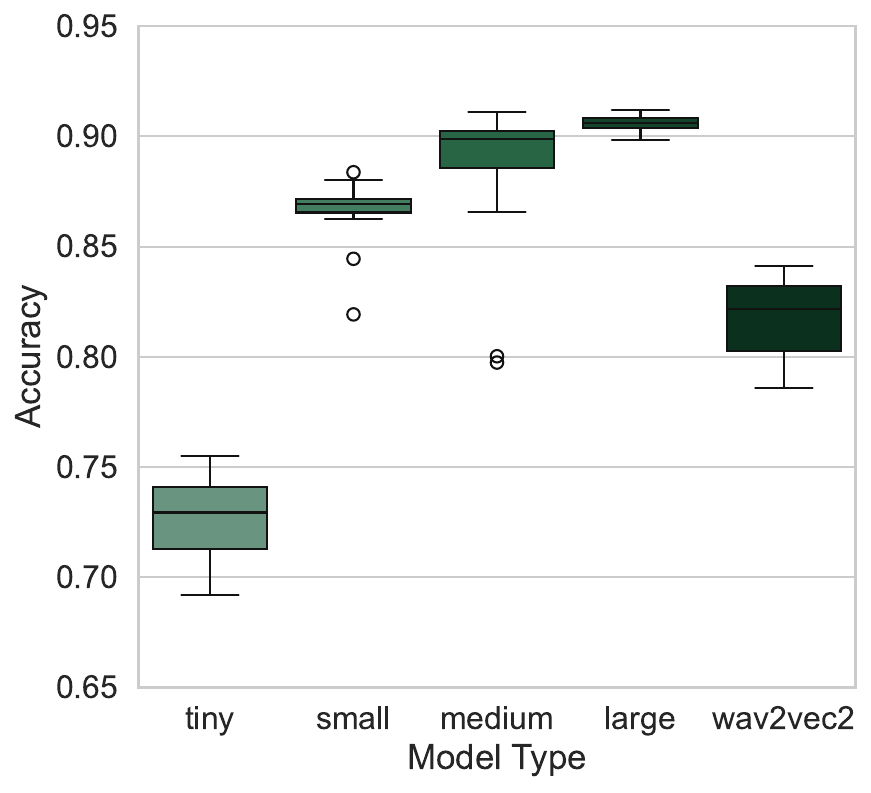}
            \caption{Test acc.~on questionhood classification.}
            \label{fig:audio-only-question-acc}
        \end{subfigure}
        \begin{subfigure}{\textwidth}
            \centering
            \includegraphics[width=\textwidth]{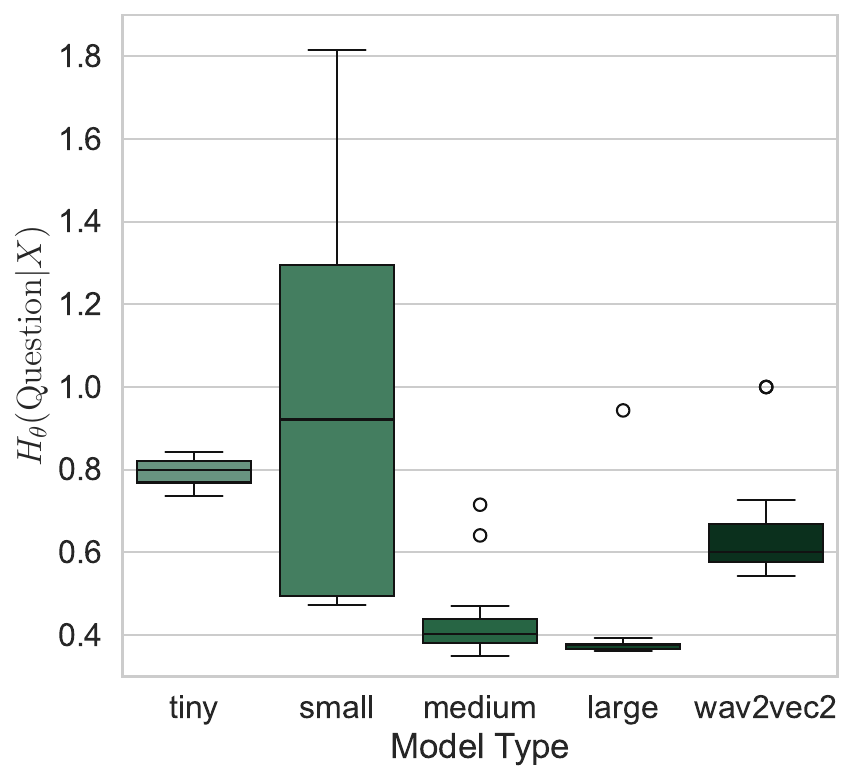}
            \caption{Test loss on questionhood classification.}
            \label{fig:audio-only-question-loss}
        \end{subfigure}
    \end{minipage}
    
    \caption{\texttt{audio-only} (Whisper + wav2vec) performance on sarcasm (left), affect (middle), and questionhood (right) classification}
    \label{fig:audio-only-acc-loss-comparison}
\end{figure*}

\end{document}

%% file: macros.tex
\newcommand{\rv}[1]{\mathrm{#1}}

\newcommand{\support}[1]{\mathcal{#1}}
\newcommand{\rvs}[1]{\boldsymbol{#1}}
\newcommand{\rvsprime}[1]{\rvs{#1}'}

\newcommand{\defvarmacros}[5]{%
  \expandafter\newcommand\csname #5\endcsname{\textcolor{#1}{\ensuremath{\rv{#3}}}\xspace}%
  \expandafter\newcommand\csname #5support\endcsname{\textcolor{#1}{\ensuremath{\support{#3}}}\xspace}%
  \expandafter\newcommand\csname #4\endcsname{\textcolor{#1}{\ensuremath{\rv{#2}}}\xspace}%
  \expandafter\newcommand\csname #5s\endcsname{\textcolor{#1}{\ensuremath{\rvs{#3}}}\xspace}%
  \expandafter\newcommand\csname #5sprime\endcsname{\textcolor{#1}{\ensuremath{\rvsprime{#3}}}\xspace}%
  \expandafter\newcommand\csname #4s\endcsname{\textcolor{#1}{\ensuremath{\rvs{#2}}}\xspace}%
  \expandafter\newcommand\csname #4sprime\endcsname{\textcolor{#1}{\ensuremath{\rvsprime{#2}}}\xspace}%
}

\definecolor{AccentRed}{RGB}{196, 65, 0}
\definecolor{AccentOrange}{RGB}{230, 122, 0}
\definecolor{AccentYellow}{RGB}{255, 255, 184}
\definecolor{AccentGreen}{RGB}{70,160,73}
\definecolor{AccentBlue}{RGB}{0,122,204}
\definecolor{AccentPurple}{RGB}{108, 0, 150}
\definecolor{AccentGray}{RGB}{160,160,160}
\definecolor{AccentLime}{RGB}{162, 207, 0}

\defvarmacros{AccentOrange}{y}{Y}{feat}{Feat}
\defvarmacros{AccentGreen}{x}{X}{channel}{Channel}
\defvarmacros{AccentBlue}{p}{P}{prosody}{Prosody}
\defvarmacros{AccentPurple}{a}{A}{audio}{Audio}
\defvarmacros{AccentRed}{t}{T}{textvar}{Text}

\newcommand{\othercommand}[2]{\newcommand{#1}{\textcolor{AccentGray}{\ensuremath{#2}}\xspace}}
\othercommand{\mi}{\mathrm{MI}}
\othercommand{\entropy}{\mathrm{H}}
\othercommand{\model}{\mathrm{q}_{\vtheta}}
\othercommand{\xent}{\entropy_{\vtheta}}
\othercommand{\KL}{\mathrm{D}_{KL}}
\othercommand{\prob}{p}

\newcommand{\newterm}[1]{\textbf{#1}}
\newcommand{\defn}[1]{\textbf{#1}}

\othercommand{\dataset}{\mathcal{D}}
\othercommand{\datasettrain}{\ensuremath{\dataset_{\mathrm{train}}}\xspace}
\othercommand{\datasettest}{\ensuremath{\dataset_{\mathrm{test}}}\xspace}
\othercommand{\ntest}{N}
\othercommand{\indicator}{\mathbb{1}}

\othercommand{\vtheta}{\boldsymbol{\theta}}
\othercommand{\ptheta}{\ensuremath{p_{\vtheta}}\xspace}
\othercommand{\ftheta}{f_{\vtheta}}
\othercommand{\vphi}{\boldsymbol{\phi}}
\othercommand{\preddist}{\mathcal{Z}}
\othercommand{\kernelfunc}{\mathrm{K}}

\newcommand{\todo}[1][]{}
\newcommand{\tocite}[1][]{}

%% file: venn_diagram_macros.tex
\usepackage{tikz}
\usetikzlibrary{patterns}

\newcommand{\legendCircleOne}{\tikz[baseline=-0.5ex]{\draw[ultra thin, fill=white] (0,0.05) circle (0.4em); \node[text=black, font=\scriptsize] at (0,0.05) {\textbf{1}};}\xspace}
\newcommand{\legendCircleTwo}{\tikz[baseline=-0.5ex]{\draw[ultra thin, fill=white] (0,0.05) circle (0.4em); \node[text=black, font=\scriptsize] at (0,0.05) {\textbf{2}};}\xspace}
\newcommand{\legendCircleThree}{\tikz[baseline=-0.5ex]{\draw[ultra thin, fill=white] (0,0.05) circle (0.4em); \node[text=black, font=\scriptsize] at (0,0.05) {\textbf{3}};}\xspace}
\newcommand{\legendCircleFour}{\tikz[baseline=-0.5ex]{\draw[ultra thin, fill=white] (0,0.05) circle (0.4em); \node[text=black, font=\scriptsize] at (0,0.05) {\textbf{4}};}\xspace}

\newcommand{\legendCircleEight}{\tikz[baseline=-0.5ex]{\draw[ultra thin, fill=white] (0,0.05) circle (0.4em); \node[text=black, font=\scriptsize] at (0,0.05) {\textbf{8}};}\xspace}
\newcommand{\legendCircleNine}{\tikz[baseline=-0.5ex]{\draw[ultra thin, fill=white] (0,0.05) circle (0.4em); \node[text=black, font=\scriptsize] at (0,0.05) {\textbf{9}};}\xspace}
\newcommand{\legendCircleTen}{\tikz[baseline=-0.5ex]{\draw[ultra thin, fill=white] (0,0.05) circle (0.4em); \node[text=black, font=\scriptsize, scale=0.7] at (0,0.05) {\textbf{10}};}\xspace}



%% file: venn_diagram.tex
\begin{tikzpicture}[scale=0.6]
    \colorlet{fillcolor}{gray!50}
    \colorlet{fillcolor2}{black}
    
    \pgfdeclarepatternformonly{dashedLines}
    {\pgfqpoint{-1pt}{-1pt}}
    {\pgfqpoint{10pt}{10pt}}
    {\pgfqpoint{9pt}{9pt}}
    {
      \pgfsetdash{{3pt}{3pt}}{0pt}  
      \pgfpathmoveto{\pgfqpoint{0pt}{0pt}}
      \pgfpathlineto{\pgfqpoint{9pt}{9pt}}
      \pgfusepath{stroke}
    }

    \begin{scope}[scale=1.2, xshift=1cm]
        \fill[pattern={Lines[angle=90, distance=8pt, line width=0.6pt]}, pattern color=fillcolor] (1,-0.2) ellipse (3.8cm and 2.6cm);
        \draw[thick] (1,-0.2) ellipse (3.8cm and 2.6cm);
    
        \begin{scope}
            \fill[pattern={dashedLines[angle=45, distance=8pt, line width=0.6pt]}, pattern color=fillcolor] (0,0) circle (2cm);
            \fill[pattern={Lines[angle=-45, distance=8pt, line width=0.6pt]}, pattern color=fillcolor] (2,0) circle (2cm);
            \fill[pattern={Lines[angle=0, distance=8pt, line width=0.6pt]}, pattern color=fillcolor] (1,2) circle (2cm);
        \end{scope}
    
        \draw[thick, dashed] (0,0) circle (2cm);
        \draw[thick] (2,0) circle (2cm);
        \draw[thick] (1,2) circle (2cm);
        
        \node at (1,-2.8) [fill=white] {$\entropy(\Audio)$};
        \node at (-1.4,-1.1) [fill=white] {$\entropy(\Prosody)$};
        \node at (3.4,-1.1) [fill=white] {$\entropy(\Text)$};
        \node at (1,4) [fill=white] {$\entropy(\Feat)$};
    \end{scope}
    
    \begin{scope}[shift={(-3.5,-4)}]
        
        \begin{scope}[shift={(4,-1)}, scale=0.25]
            \draw[thick] (1,-0.5) ellipse (3.8cm and 2.8cm);
            \draw[thick, dashed] (0,0) circle (2cm);
            \draw[thick] (2,0) circle (2cm);
            \draw[thick] (1,2) circle (2cm);
            
            \begin{scope}
                \clip (2,0) circle (2cm);
                \clip (0,0) circle (2cm);
                \fill[fillcolor2] (-4,-4) rectangle (8,8);
            \end{scope}
        \end{scope}
        \node[anchor=west, scale=0.8] at (0.3,-1.1) {$\mi(\Text;\Prosody)$};
        \draw[thick, fill=white] (0,-1) circle (0.35cm);
        \node[text=black] at (0,-1) {\textbf{1}};


        \begin{scope}[shift={(4,-3)}, scale=0.25]
            \draw[thick] (1,-0.5) ellipse (3.8cm and 2.8cm);
            \draw[thick, dashed] (0,0) circle (2cm);
            \draw[thick] (2,0) circle (2cm);
            \draw[thick] (1,2) circle (2cm);
            \begin{scope}    
                \clip (1,2) circle (2cm);
                \clip (0,0) circle (2cm);
                \clip (2,0) circle (2cm) (-4,-4) rectangle (8,8);
                \fill[fillcolor2] (-4,-4) rectangle (8,8);
            \end{scope}
        \end{scope}
        \node[anchor=west, scale=0.8] at (0.3,-3.1) {$\mi(\Feat;\Prosody|\Text)$};
        \draw[thick, fill=white] (0,-3) circle (0.35cm);
        \node[text=black] at (0,-3) {\textbf{2}};

        \begin{scope}[shift={(4,-5)}, scale=0.25]
            \draw[thick] (1,-0.5) ellipse (3.8cm and 2.8cm);
            \draw[thick, dashed] (0,0) circle (2cm);
            \draw[thick] (2,0) circle (2cm);
            \draw[thick] (1,2) circle (2cm);
            \begin{scope}
                \clip (2,0) circle (2cm);
                \clip (1,2) circle (2cm);
                \fill[fillcolor2] (-2,-2) rectangle (4,4);
            \end{scope}
        \end{scope}
        \node[anchor=west, scale=0.8] at (0.3,-5.1) {$\mi(\Feat;\Text)$};
        \draw[thick, fill=white] (0,-5) circle (0.35cm);
        \node[text=black] at (0,-5) {\textbf{3}};

        \begin{scope}[shift={(4,-7)}, scale=0.25]
            \draw[thick] (1,-0.5) ellipse (3.8cm and 2.8cm);
            \draw[thick, dashed] (0,0) circle (2cm);
            \draw[thick] (2,0) circle (2cm);
            \draw[thick] (1,2) circle (2cm);
            \begin{scope}
                \clip (1,2) circle (2cm);
                \clip (1,-0.5) ellipse (3.8cm and 2.8cm);
                \fill[fillcolor2] (-2,-2) rectangle (4,4);
            \end{scope}
        \end{scope}
        \node[anchor=west, scale=0.8] at (0.3,-7.1) {$\mi(\Feat;\Audio)$};
        \draw[thick, fill=white] (0,-7) circle (0.35cm);
        \node[text=black] at (0,-7) {\textbf{4}};

        \begin{scope}[shift={(4,-9)}, scale=0.25]
            \draw[thick] (1,-0.5) ellipse (3.8cm and 2.8cm);
            \draw[thick, dashed] (0,0) circle (2cm);
            \draw[thick] (2,0) circle (2cm);
            \draw[thick] (1,2) circle (2cm);
            
            \begin{scope}
                \clip (1,2) circle (2cm);
                \clip (0,0) circle (2cm);
                \clip (2,0) circle (2cm);
                
                \fill[fillcolor2] (-4,-4) rectangle (8,8);
            \end{scope}
        \end{scope}
        \node[anchor=west, scale=0.8] at (0.3,-9.1) {$\mi(\Feat;\Prosody;\Text)$};
        \draw[thick, fill=white] (0,-9) circle (0.35cm);
        \node[text=black] at (0,-9) {\textbf{5}};
        
        \begin{scope}[shift={(11,-1)}, scale=0.25]
            \draw[thick] (1,-0.5) ellipse (3.8cm and 2.8cm);
            \draw[thick, dashed] (0,0) circle (2cm);
            \draw[thick] (2,0) circle (2cm);
            \draw[thick] (1,2) circle (2cm);
            
            \begin{scope}
                \clip (1,2) circle (2cm);
                \clip (0,0) circle (2cm) (-4,-4) rectangle (8,8);
                \fill[fillcolor2] (-4,-4) rectangle (8,8);
            \end{scope}
        \end{scope}
        \node[anchor=west, scale=0.8] at (6.9,-1.1) {$\entropy(\Feat|\Prosody)$};
        \draw[thick, fill=white] (6.6,-1) circle (0.35cm);
        \node[text=black] at (6.6,-1) {\textbf{6}};

        \begin{scope}[shift={(11,-3)}, scale=0.25]
            \draw[thick] (1,-0.5) ellipse (3.8cm and 2.8cm);
            \draw[thick, dashed] (0,0) circle (2cm);
            \draw[thick] (2,0) circle (2cm);
            \draw[thick] (1,2) circle (2cm);
            
            \begin{scope}
                \clip (1,2) circle (2cm);
                \clip (2,0) circle (2cm) (-4,-4) rectangle (8,8);
                \fill[fillcolor2] (-4,-4) rectangle (8,8);
            \end{scope}
        \end{scope}
        \node[anchor=west, scale=0.8] at (6.9,-3.1) {$\entropy(\Feat|\Text)$};
        \draw[thick, fill=white] (6.6,-3) circle (0.35cm);
        \node[text=black] at (6.6,-3) {\textbf{7}};

        \begin{scope}[shift={(11,-5)}, scale=0.25]
            \draw[thick] (1,-0.5) ellipse (3.8cm and 2.8cm);
            \draw[thick, dashed] (0,0) circle (2cm);
            \draw[thick] (2,0) circle (2cm);
            \draw[thick] (1,2) circle (2cm);
            
            \begin{scope}
                \clip (1,2) circle (2cm);
                \clip (1,-0.5) ellipse (3.8cm and 2.8cm) (-4,-4) rectangle (8,8);
                \fill[fillcolor2] (-4,-4) rectangle (8,8);
            \end{scope}
        \end{scope}
        \node[anchor=west, scale=0.8] at (6.9,-5.1) {$\entropy(\Feat|\Audio)$};
        \draw[thick, fill=white] (6.6,-5) circle (0.35cm);
        \node[text=black] at (6.6,-5) {\textbf{8}};

        \begin{scope}[shift={(11,-7)}, scale=0.25]
            \draw[thick] (1,-0.5) ellipse (3.8cm and 2.8cm);
            \draw[thick, dashed] (0,0) circle (2cm);
            \draw[thick] (2,0) circle (2cm);
            \draw[thick] (1,2) circle (2cm);
            \begin{scope}
                \begin{scope}
                    \clip (1,2) circle (2cm);
                    \clip (1,-0.5) ellipse (3.8cm and 2.8cm);
                    \begin{scope}
                        \clip (-4,-4) rectangle (8,8);
                        \clip (-4,-4) rectangle (8,8) (-2,0) rectangle (6,-4) (2,0) circle (2cm);
                        \fill[fillcolor2] (-4,-4) rectangle (8,8);
                    \end{scope}
                \end{scope}
            \end{scope}
        \end{scope}
        \node[anchor=west, scale=0.8] at (6.9,-7.1) {$\mi(\Feat;\Audio|\Text)$};
        \draw[thick, fill=white] (6.6,-7) circle (0.35cm);
        \node[text=black] at (6.6,-7) {\textbf{9}};

        \begin{scope}[shift={(11,-9)}, scale=0.25]
            \draw[thick] (1,-0.5) ellipse (3.8cm and 2.8cm);
            \draw[thick, dashed] (0,0) circle (2cm);
            \draw[thick] (2,0) circle (2cm);
            \draw[thick] (1,2) circle (2cm);
            
            \begin{scope}
                \clip (1,2) circle (2cm);
                \clip (1,-0.5) ellipse (3.8cm and 2.8cm);
                \clip (0,0) circle (2cm) (-4,-4) rectangle (8,8);
                \clip (2,0) circle (2cm) (-4,-4) rectangle (8,8);
                
                \fill[fillcolor2] (-4,-4) rectangle (8,8);
            \end{scope}
        \end{scope}
        \node[anchor=west, scale=0.8] at (6.9,-9.1) {$\mi(\Feat;\Audio|\Text,\Prosody)$};
        \draw[thick, fill=white] (6.6,-9) circle (0.35cm);
        \node[text=black,scale=0.8] at (6.6,-9) {\textbf{10}};
        
    \end{scope}
\end{tikzpicture}